\documentclass[lettersize,journal]{IEEEtran}
\usepackage{graphicx} 
\usepackage{algorithm}
\usepackage{algorithmic}
\usepackage{amsmath}
\usepackage{amssymb}
\usepackage{pifont}

\usepackage{tikz}
\usepackage{ifthen}

\usepackage{multirow}
\usepackage{booktabs}
\usepackage{multirow}
\usepackage{xcolor}
\usepackage{graphicx}

\title{Remote Sensing Retrieval-Augmented Generation: Bridging Remote Sensing Imagery and Comprehensive Knowledge with a Multi-Modal Dataset and Retrieval-Augmented Generation Model}

\author{
Congcong Wen \IEEEmembership{Member, IEEE}, Yiting Lin, Xiaokang Qu, Nan Li, Yong Liao, Xiang Li and  Hui Lin \\

\thanks{This work was partially supported by the Beijing Nova Program (Grant No. 2024124) and the National Natural Science Foundation of China (Grant Nos. U24B20177, U25B2075, and U2541212). (Congcong Wen and Yiting Lin contributed equally to this work). \textit{(Corresponding author: Hui Lin)}.}

\thanks{Congcong Wen, Yiting Lin, Xiaokang Qu and Yong Liao are with School of Cyber Science and Technology, University of Science and Technology of China, Anhui, 230026, China. Congcong Wen is also with China Academy of Electronics and Information Technology, Beijing 100846, China. (e-mail: wencc1208@gmail.com, linyiting@mail.ustc.edu.cn, xkqu@mail.ustc.edu.cn and yliao@ustc.edu.cn)}

\thanks{Nan Li and Hui Lin are with China Academy of Electronics and Information Technology, Beijing 100846, China. (e-mail: linhui@whu.edu.cn and nli2014@lzu.edu.cn)}

\thanks{Xiang Li is with the School of Computer Science at the University of Bristol, UK. (e-mail: xiangli92@ieee.org)}

}

\newcommand{\revise}[1]{\textcolor{black}{#1}}
\newcommand{\rrevise}[1]{\textcolor{black}{#1}}

\begin{document}

\maketitle

\begin{abstract}
\revise{Recent progress in Vision-Language Models (VLMs) has demonstrated impressive capabilities across a variety of tasks in the natural image domain. Motivated by these advancements, the remote sensing community has begun to adopt VLMs for remote sensing vision-language tasks, including image captioning, scene understanding, and visual question answering (VQA). However, existing remote sensing VLMs typically rely on closed-set scene understanding and focus on generic scene descriptions, yet lack the ability to incorporate external knowledge. This limitation hinders their capacity for semantic reasoning over complex or context-dependent queries that involve domain-specific or world knowledge. To address these challenges, we first introduced a multimodal Remote Sensing World Knowledge (RSWK) dataset, which comprises high-resolution satellite imagery and detailed textual descriptions for 14,820 well-known landmarks across 16 categories from 184 countries, integrating both remote sensing domain knowledge and broader world knowledge. Building upon this dataset, we proposed a novel Remote Sensing Retrieval-Augmented Generation (RS-RAG) framework, which consists of two key components. The Multi-Modal Knowledge Vector Database Construction module encodes remote sensing imagery and associated textual knowledge into a unified vector space. The Knowledge Retrieval and Response Generation module retrieves and re-ranks relevant knowledge based on image and/or text queries, and incorporates the retrieved content into a knowledge-augmented prompt to guide the VLM in producing contextually grounded responses. We validate the effectiveness of RS-RAG on our designed benchmark covering three vision-language tasks: image captioning, image classification, and VQA, where RS-RAG significantly outperformed state-of-the-art baselines. By bridging remote sensing imagery and comprehensive knowledge, RS-RAG empowers remote sensing VLMs with enhanced contextual reasoning, enabling them to generate more accurate, informative, and semantically grounded outputs across a wide range of tasks.}

\end{abstract}


\section{Introduction}

Remote sensing imagery, as a critical source of information for Earth observation and monitoring, plays an essential role in urban planning~\cite{han2024autoencoding}, agricultural assessment~\cite{liu2020improved}, and environmental protection~\cite{wen2019novel}. However, as remote sensing technology advances, the scale and complexity of imagery data have rapidly increased, making it increasingly difficult for traditional manual analysis or image processing methods to meet practical demands. Deep learning methods~\cite{zhu2017deep,lin2025generalization} have significantly improved the accuracy and efficiency of tasks like classification, segmentation, and object detection by automatically extracting features from vast amounts of remote sensing data. While these methods have made notable progress, most deep learning models rely predominantly on single-modal visual information, lacking deep semantic understanding of image content. This limitation results in reduced generalization and adaptability, particularly for tasks that require in-depth semantic analysis and comprehensive scene understanding.

The emergence of Vision-Language Models (VLMs)~\cite{hu2025rsgpt,bazi2024rs,pang2024h2rsvlm,zhang2024earthgpt,kuckreja2024geochat,zhan2025skyeyegpt} offers a novel solution for the semantic analysis of remote sensing data. By leveraging multimodal fusion techniques, VLMs combine visual features with language information to automatically generate descriptive insights for remote sensing imagery. This semantic enhancement improves image classification and object detection performance while enabling the transformation of recognition results into natural language descriptions, making them more interpretable and accessible for various applications. Additionally, VLMs perform well even in weakly supervised or zero-shot learning scenarios, providing reliable analysis with limited labeled data and thus reducing dependency on extensive data annotations. This cross-modal integration not only enhances the model’s cognitive ability to interpret remote sensing imagery but also facilitates detailed semantic descriptions of complex scenes, paving the way for broader intelligent applications in remote sensing. However, existing remote sensing VLMs primarily focus on identifying image features and providing basic scene descriptions, lacking a deeper background understanding of the objects within the images. Particularly when it comes to rich semantic information requiring remote sensing domain expertise or other general world knowledge, such as historical, cultural, and social contexts, these \revise{VLMs} often struggle to provide comprehensive contextual support.

\revise{To address this issue, we first introduce the Remote Sensing World Knowledge (RSWK) dataset, a multimodal benchmark that includes high-resolution imagery and natural language descriptions for over 14,000 well-known locations across 16 categories in 184 countries worldwide.} Unlike most existing remote sensing vision-language datasets that only provide basic descriptions of current scenes, our RSWK dataset includes richer remote sensing domain expertise and world knowledge about the objects within these scenes. For instance, from a remote sensing perspective, it provides information on surface reflectance, spectral indices, and atmospheric conditions. From a world knowledge perspective, it includes historical background, cultural significance, construction period, and major events. This combination of remote sensing expertise and world knowledge not only enhances the RSWK dataset's utility for visual analysis of remote sensing images, but also provides the model with deeper semantic context, overcoming the limitations of traditional datasets and enabling remote sensing VLMs to perform more complex cognitive tasks. Furthermore, our dataset incorporates historical, cultural, and social backgrounds from various countries and regions, allowing \revise{VLMs} to be trained across diverse geographical and cultural contexts, thereby improving their generalization ability and understanding of different cultural settings.

\begin{figure*}[t]
    \centering
    \includegraphics[width=1.0\linewidth]{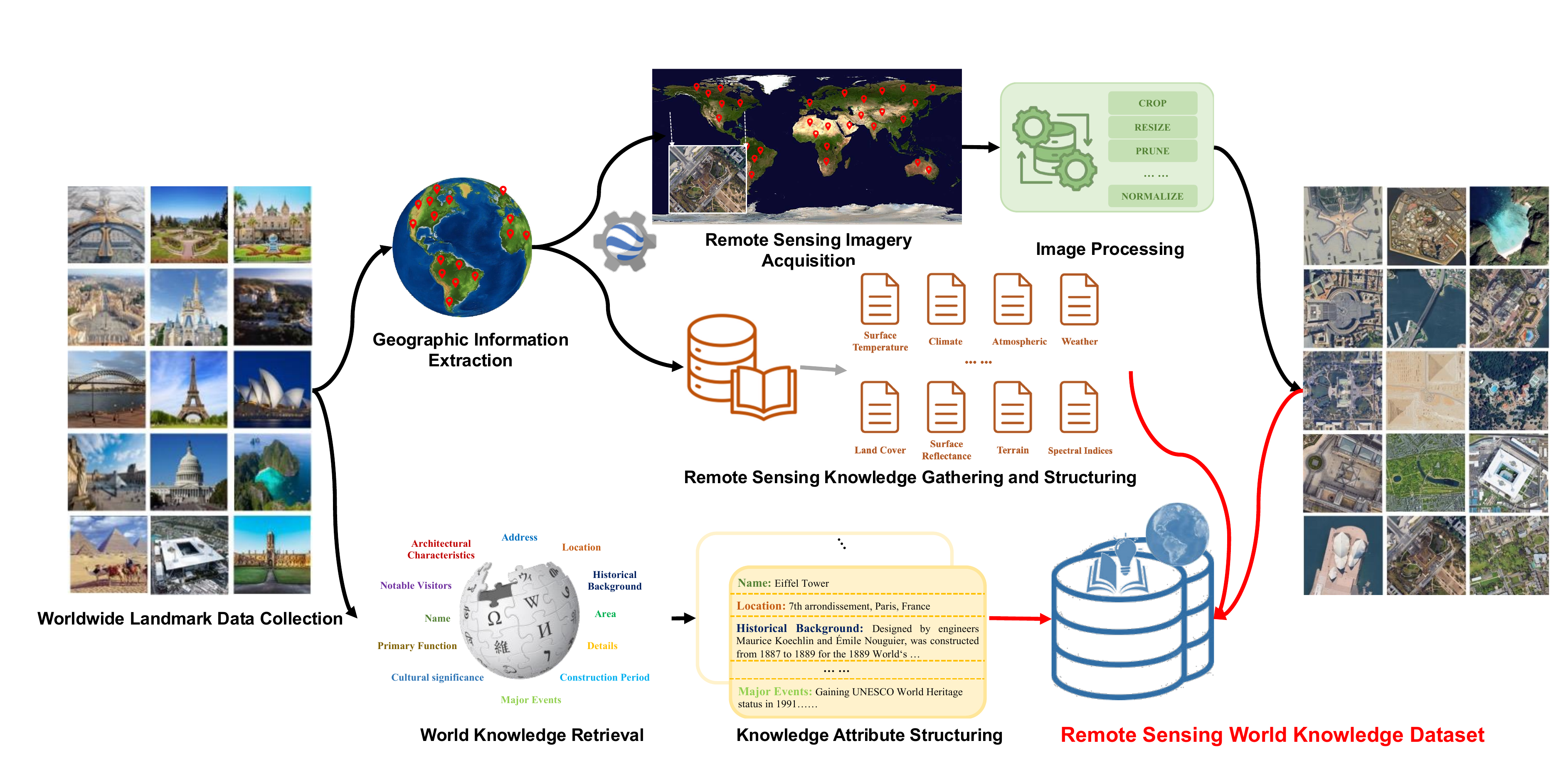}
    \caption{The construction process of the Remote Sensing World Knowledge (RSWK) dataset begins with collecting landmark data from around the world, followed by extracting geographic information to pinpoint precise location coordinates. Using these coordinates, remote sensing images are acquired, which are further standardized through image processing techniques. Corresponding remote sensing expert knowledge, such as surface temperature, climate, atmospheric conditions, and spectral \revise{indices}, is also included. Additionally, world knowledge is retrieved from online resources, providing detailed background information about the landmarks, including historical context, cultural significance, and major events. This combined information is structured into organized attributes to align image and text data, forming a final multimodal dataset. The resulting RSWK dataset integrates high-resolution images with extensive remote sensing and world knowledge, enabling advanced semantic understanding in remote sensing applications.}
    \label{fig:rswk}
\end{figure*}

Furthermore, to effectively leverage the comprehensive information provided by the RSWK dataset, we propose the Remote Sensing Retrieval-Augmented Generation (RS-RAG) model. RS-RAG is designed to enhance the capacity of \revise{VLMs} to generate contextually enriched and knowledge-grounded responses for remote sensing imagery. It operates by integrating external knowledge, both domain-specific and general world knowledge, retrieved from a multimodal knowledge base constructed using the RSWK dataset. The model consists of two main components: (1) the Multi-Modal Knowledge Vector Database Construction module, which encodes satellite imagery and textual descriptions into a shared embedding space using unified image and text encoders to enable efficient cross-modal retrieval; and (2) the Knowledge Retrieval and Response Generation module, which retrieves top-ranked knowledge entries based on image and text queries, re-ranks them via a fused similarity score, and incorporates the selected content into a knowledge-augmented prompt. This prompt is then passed to the vision-language model, enabling it to generate responses that go beyond semantic understanding and reflect deeper background knowledge. By coupling visual input with relevant contextual information, RS-RAG significantly improves the interpretability and accuracy of vision-language outputs, particularly for complex queries involving geospatial, historical, or environmental reasoning.

\revise{To assess the effectiveness of the proposed model, we construct a comprehensive benchmark and evaluate performance across three representative tasks: image captioning, image classification, and visual question answering (VQA). To better leverage the rich semantics of the dataset, we further refine the VQA task into three distinct subtypes that target: (1) comprehensive reasoning, (2) domain-specific remote sensing understanding, and (3) general world knowledge interpretation.} Results on these tasks demonstrate the model’s ability to produce accurate, context-rich descriptions, deliver semantically informed scene classifications, and provide precise answers to knowledge-intensive queries by leveraging both visual and textual modalities. Through these findings, RS-RAG demonstrates its potential to substantially advance the capabilities of \revise{VLMs} in remote sensing, effectively bridging the gap between imagery and comprehensive contextual knowledge. Our main contributions are summarized as follows:

\begin{itemize}
    \item \revise{We construct the Remote Sensing World Knowledge (RSWK) dataset, a large-scale multimodal dataset containing 14,820 high-resolution remote sensing images and rich textual descriptions of globally recognized landmarks across 16 categories from 184 countries. The descriptions incorporate both domain-specific remote sensing knowledge and general world knowledge.}

    \item We propose RS-RAG, a novel Retrieval-Augmented Generation framework tailored for remote sensing vision-language tasks. RS-RAG retrieves semantically relevant knowledge from a multimodal vector database and integrates it with the input via knowledge-conditioned prompt construction, significantly enhancing contextual reasoning capabilities.

    \item \revise{We design a benchmark spanning image captioning, classification, and VQA, with VQA further divided into three subtypes to assess comprehensive reasoning, remote sensing expertise, and world knowledge. This setup facilitates systematic evaluation of both semantic perception and knowledge-intensive reasoning.}

    \item Extensive experiments demonstrate that RS-RAG consistently outperforms state-of-the-art VLMs across all tasks, particularly on queries requiring external world knowledge. These results highlight the effectiveness of RS-RAG in bridging remote sensing imagery with structured knowledge, and point to promising future directions for research on remote sensing vision-language models.
\end{itemize}


\section{Related Work}

\subsection{Remote Sensing Multimodal Datasets}
Several multimodal datasets~\cite{lobry2020rsvqa,zheng2021mutual,sun2022visual,al2022open,zhan2023rsvg} have been developed to bridge the gap between vision and language in the remote sensing domain. UCM Captions~\cite{ucm_sydney_caption}, Sydney Captions~\cite{ucm_sydney_caption}, and RSICD~\cite{rsicd} are among the earliest datasets that provide textual descriptions for remote sensing images. Each image in these datasets is paired with five relatively simple human-written sentences, offering only a basic level of semantic information. To enhance the quality and richness of textual annotations, RSGPT~\cite{hu2025rsgpt} recently introduced RSICap, a high-quality image captioning dataset with detailed human-annotated descriptions of aerial scenes. RSICap serves as a valuable resource for fine-tuning and developing domain-specific vision-language models in remote sensing. Beyond manual annotation, researchers have explored the construction of large-scale datasets using automatic or hybrid approaches applied to existing data sources. For instance, RS5M~\cite{zhang2024rs5m} was created by aggregating 11 publicly available image–text paired datasets along with three large-scale class-level labeled datasets. Captions were generated using BLIP-2, resulting in a diverse and large-scale dataset suitable for training foundational multimodal models. Similarly, RemoteCLIP~\cite{remoteclip} compiled a large-scale dataset by integrating 10 object detection datasets, 4 semantic segmentation datasets, and 3 remote sensing image–text datasets, enabling contrastive pretraining for cross-modal alignment. In addition, GeoChat~\cite{geochat} introduced the RS Multimodal Instruction Dataset, which incorporates heterogeneous data sources, including three object detection datasets, one scene classification dataset, and a visual question answering dataset focused on flood detection. \revise{Recent datasets such as SkyScript~\cite{wang2024skyscript}, LhrsAlign~\cite{muhtar2024lhrs}, and ChatEarthNet~\cite{yuan2024chatearthnet} leverage external geospatial sources, such as OpenStreetMap, along with geo-coordinates to construct large-scale, globally distributed multimodal benchmarks.} More recently, FedRSCLIP~\cite{lin2025fedrsclip} proposed a new multimodal remote sensing dataset specifically designed for federated learning scenarios, further expanding the applicability of vision-language research in distributed settings. These efforts collectively advance the development of remote sensing VLMs by providing diverse, rich, and large-scale multimodal data resources tailored to various downstream tasks.

\subsection{Remote Sensing Multimodal Model}
With the growing availability of remote sensing multimodal datasets, a number of vision-language models~\cite{hu2025rsgpt,bazi2024rs,pang2024h2rsvlm,zhang2024earthgpt,kuckreja2024geochat,zhan2025skyeyegpt} have been proposed to enhance the understanding and interpretation of aerial imagery through natural language. \cite{li2024vision} provides the first comprehensive review of vision-language models in remote sensing, systematically summarizing tasks, datasets, and methods, and identifying key challenges and future directions. RSCLIP~\cite{rs-clip} is a pioneering vision-language model designed for remote sensing scene classification, which leverages contrastive vision-language supervision and incorporates a pseudo-labeling technique along with a curriculum learning strategy to improve zero-shot classification performance through multi-stage fine-tuning. RSGPT~\cite{hu2025rsgpt} represents one of the earliest attempts to build a generative pretrained model for remote sensing. It fine-tunes only the Q-Former and a linear projection layer on the proposed RSICap dataset, achieving notable improvements in both image captioning and visual question answering tasks. Similarly, GeoChat~\cite{geochat} adapts the LLaVA-1.5 architecture and fine-tunes it on its proposed remote sensing multimodal instruction-following dataset, offering multi-task conversational capabilities grounded in high-resolution satellite imagery. In addition, to address the common issue of hallucination in remote sensing vision-language models, a Helpful and Honest Remote Sensing Vision-Language Model~\cite{pang2024h2rsvlm}, named H2RSVLM, is proposed and fine-tuned on the RSSA dataset, the first dataset specifically designed to enhance self-awareness in remote sensing VLMs. More recently, RS-MoE~\cite{lin2025rs} was proposed as the first Mixture-of-Experts-based VLM tailored for remote sensing. It features a novel instruction router that dynamically dispatches tasks to multiple lightweight expert LLMs, allowing each expert to specialize in a specific subset of tasks.

\begin{figure*}[t]
    \centering
    \includegraphics[width=1.0\linewidth]{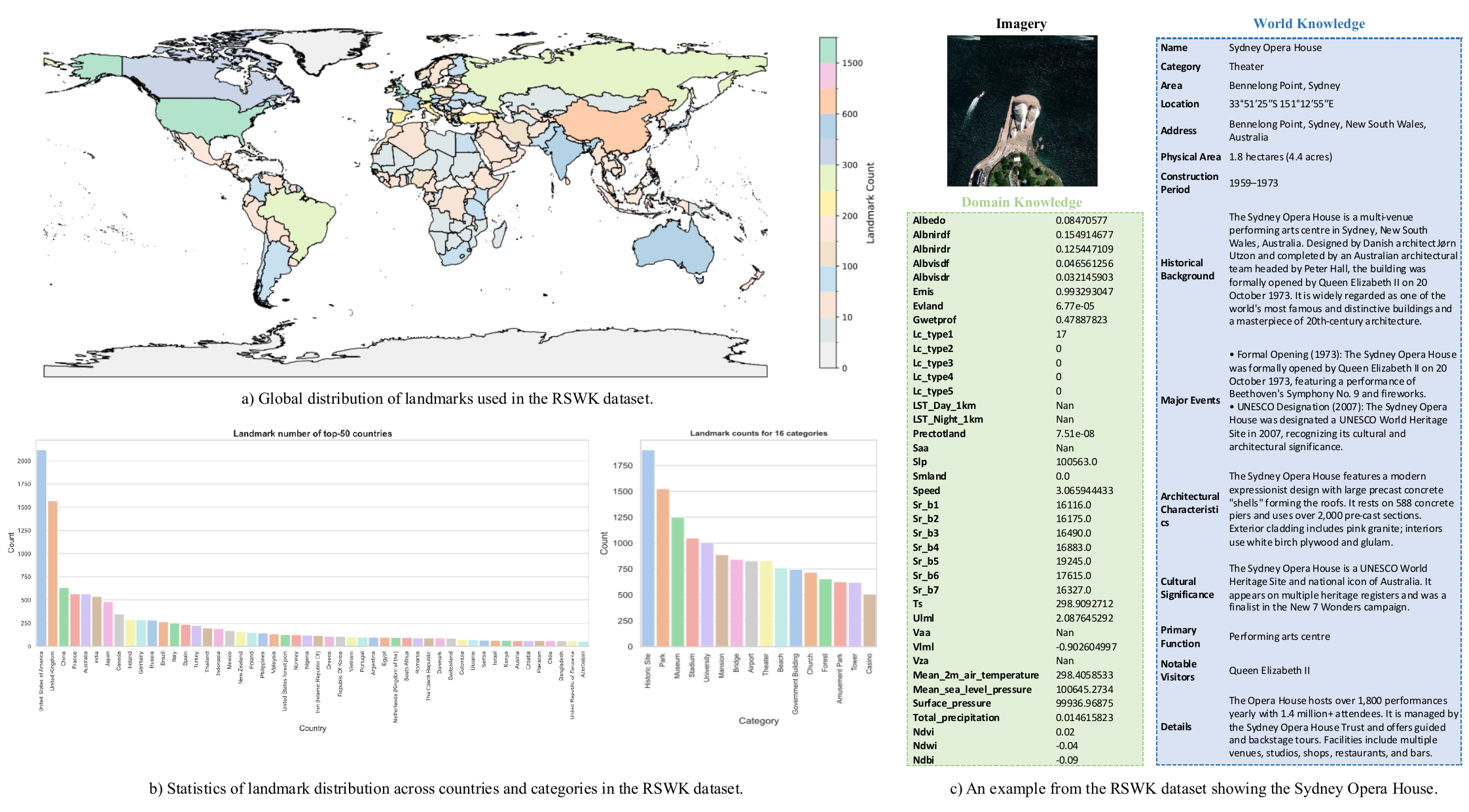}
    \caption{Overview of the RSWK dataset. (a) Global distribution of landmarks used in the dataset, with color indicating the number of landmarks per country. (b) Statistical summaries of landmark counts across the top \revise{50} countries (left) and \revise{all 16} categories (right). (c) A specific example from the RSWK dataset, showcasing the Sydney Opera House, including its satellite imagery, remote sensing domain knowledge, and structured world knowledge.}
    \label{fig:rswk_overview}
\end{figure*}

{\color{black}
\subsection{Retrieval-Augmented Generationl} 

To address the limitations of VLMs in handling knowledge-intensive or domain-specific tasks, Retrieval-Augmented Generation (RAG)~\cite{zheng2025retrieval} has emerged as a promising solution for enhancing the accuracy and reliability of generative models by incorporating information from external, relevant knowledge sources. By retrieving and integrating contextual knowledge during inference, RAG significantly improves the reasoning ability and factual consistency of large models. In recent years, researchers have explored the integration of RAG into a variety of domains, including natural language processing\cite{fan2024survey}, natural image understanding\cite{long2022retrieval}, and video analysis~\cite{jeong2025videorag}. However, the application of RAG in the remote sensing domain remains limited. More recently, RAGCap~\cite{bazi2025ragcap} proposes a fine-tuning-free method for remote sensing image captioning, where relevant image-caption pairs are retrieved from a training set to guide generation. Similarly, ImageRAG~\cite{zhang2024imagerag} introduces an image-contextual retrieval mechanism based on RAG principles, which transforms ultra-high-resolution remote sensing image analysis into a long-context selection task, enabling efficient understanding with minimal training data.

}




\section{Dataset Construction}

Most existing remote sensing \revise{multimodal} datasets focus primarily on basic descriptions of objects in a scene, typically providing only information about the objects present. While these datasets serve a purpose in supporting fundamental tasks such as scene classification and image captioning, they fall short in applications requiring complex semantic understanding or contextual awareness. To address this limitation, in our paper, we propose the Remote Sensing World Knowledge (RSWK) dataset, which encompasses high-resolution remote sensing imagery of well-known locations worldwide, along with domain knowledge and world knowledge described in natural language. This dataset not only fills the gap in knowledge depth and breadth found in current remote sensing datasets but also provides a new foundation for advancing remote sensing technology toward intelligent applications. The core value of remote sensing data lies in its ability to deliver rich geographic and spatial distribution information. By integrating high-resolution imagery from across the globe with detailed domain knowledge and world knowledge, the RSWK dataset extends this value, expanding the application of remote sensing data from traditional foundational tasks to complex scenarios that require deeper semantic understanding.


\subsection{Data Collection and Processing}
Fig.~\ref{fig:rswk} illustrates the end-to-end pipeline for constructing the \revise{RSWK} dataset, which aims to bridge the gap between remote sensing domain knowledge and encyclopedic world knowledge of globally distributed landmarks. \revise{The pipeline comprises four key stages that systematically} curate and align multimodal information from global sources, integrating high-resolution remote sensing imagery, remote sensing expert knowledge, and contextual world knowledge. 

{\color{black}
\textit{1) Worldwide Landmark Data Collection:} The first stage involves generating a comprehensive list of globally recognized landmarks using GPT-4o~\cite{hurst2024gpt}. To ensure broad coverage across geographic regions and cultural contexts, we prompt the model to identify representative landmarks spanning a wide range of categories, including transportation hubs (e.g., airports), recreational areas (e.g., amusement parks, beaches), cultural institutions (e.g., museums, theaters, universities), historical sites, and natural environments (e.g., forests). 

\textit{2) Remote Sensing Imagery Acquisition}: The second stage focuses on acquiring high-resolution satellite imagery for each landmark. We first utilize the Google Geocoding Maps API to obtain precise geographic coordinates and region-level viewpoint information for each target location. To resolve potential ambiguities in place names, we implement a cross-source validation mechanism that references multiple geospatial data providers and applies regional constraints to enhance geolocation accuracy. For image acquisition, we access the ArcGIS World Imagery Tile Map Service provided by Esri, which delivers high-resolution satellite imagery in standard XYZ tile format using the Web Mercator projection. Based on the retrieved geo-coordinates, we collect imagery tiles at appropriate zoom levels ranging from 16 to 18, depending on the spatial extent of each landmark. This adaptive zoom-level strategy ensures that landmarks of varying scales (e.g., airport, forest) are fully captured within the resulting imagery. To enable efficient downloading, we implement a multithreaded queue-based retrieval system that parallelizes tile requests across locations. The downloaded tiles are then mosaicked into seamless, high-resolution scene-level images. Finally, we apply standardized normalization, cropping, and resizing operations to ensure consistent image quality, resolution, and format across the dataset.

\textit{3) Remote Sensing Knowledge Gathering and Structuring:} The third stage focuses on acquiring structured remote sensing expert knowledge using Google Earth Engine (GEE)\cite{zhao2021progress}. This knowledge spans a broad set of satellite-derived physical and environmental variables. These include surface reflectance bands, vegetation and land surface indices, as well as biophysical parameters like surface albedo, emissivity, and land cover classifications. In addition, the dataset incorporates a range of meteorological and geophysical variables obtained from authoritative products such as MERRA-2, HLS, and ERA5. These include land surface temperature, snowmelt flux, total precipitation, surface wind speed and direction, solar angles, and pressure-related measurements. To ensure temporal consistency between imagery and knowledge fields, all dynamic variables are aggregated either annually or monthly within the same year as the image acquisition.

\textit{4) World Knowledge Retrieval:} The final stage involves curating structured world knowledge to provide rich semantic context for each landmark. We leverage the Wikipedia API to extract descriptive entries for each site, encompassing diverse dimensions such as historical background, architectural characteristics, cultural significance, and major events. To ensure the relevance and quality of the extracted content, we apply a multi-step post-processing pipeline. Specifically, keyword-based filtering and entity recognition are used to discard generic or off-topic segments, while DeepSeek~\cite{liu2024deepseek} is employed to refine textual content by correcting formatting inconsistencies, improving coherence, and removing noisy or redundant information. The resulting knowledge is clean, semantically rich, and structured into predefined fields, enabling effective integration with downstream vision-language tasks that require contextual understanding and factual grounding.

}

\begin{table}[ht]
\centering
\caption{Remote Sensing Domain Knowledge Fields, Descriptions, \revise{Temporal Resolution,} and Source for Landmark in the RSWK dataset.}
\resizebox{0.5\textwidth}{!}{
\begin{tabular}{lp{5cm}ll}
\toprule
\textbf{Field Name} & \textbf{Description} & \revise{\textbf{Temporal Resolution}} & \textbf{Source} \\
\midrule
\texttt{Albedo} & Surface albedo, representing the reflectivity of the Earth's surface & \revise{Annual, Monthly} &  MERRA-2 (M2T1NXRAD) \\
\texttt{Albnirdf} & Near-infrared diffuse surface albedo & \revise{Annual, Monthly} &  MERRA-2 (M2T1NXRAD) \\
\texttt{Albnirdr} & Near-infrared direct surface albedo & \revise{Annual, Monthly} & MERRA-2 (M2T1NXRAD) \\
\texttt{Albvisdf} & Visible light diffuse albedo & \revise{Annual, Monthly} & MERRA-2 (M2T1NXRAD) \\
\texttt{Albvisdr} & Visible light direct albedo & \revise{Annual, Monthly} &  MERRA-2 (M2T1NXRAD) \\
\texttt{Emis} & Surface emissivity, related to thermal radiation & \revise{Annual, Monthly} & MERRA-2 (M2T1NXRAD) \\
\texttt{Evland} & Evaporation land (\revise{mm}) & \revise{Annual, Monthly} & MERRA-2 (M2T1NXLND) \\
\texttt{Gwetprof} & Profile soil moisture averaged over depth & \revise{Annual, Monthly} & MERRA-2 (M2T1NXLND) \\
\texttt{LC\_type1} & Land cover type (IGBP classification) & \revise{Annual} & MODIS (MCD12Q1.061) \\
\texttt{LC\_type2} & Land cover type (UMD classification) & \revise{Annual} & MODIS (MCD12Q1.061) \\
\texttt{LC\_type3} & Land cover type (LAI classification) & \revise{Annual} & MODIS (MCD12Q1.061) \\
\texttt{LC\_type4} & Land cover type (BGC classification) & \revise{Annual} & MODIS (MCD12Q1.061) \\
\texttt{LC\_type5} & Land cover type (Plant Functional Types classification) & \revise{Annual} & MODIS (MCD12Q1.061) \\
\texttt{LST\_Day\_1km} & Daytime land surface temperature (°C) & \revise{Annual, Monthly} & MODIS (MOD11A1.061) \\
\texttt{LST\_Night\_1km} & Nighttime land surface temperature (°C) & \revise{Annual, Monthly} & MODIS (MOD11A1.061) \\
\texttt{Prectotland} & Total land precipitation (\revise{mm}) & \revise{Annual, Monthly} & MERRA-2 (M2T1NXLND) \\
\texttt{Saa} & Sun Azimuth Angle (°) & \revise{Annual} & HLSL30 (HLS-2) \\
\texttt{Slp} & Sea level pressure (Pa) & \revise{Annual, Monthly} & MERRA-2 (M2T1NXSLV) \\
\texttt{Smland} & Snowmelt flux land (\revise{mm}) & \revise{Annual, Monthly} & MERRA-2 (M2T1NXLND) \\
\texttt{Speed} & Surface wind speed ($\mathrm{m}/\mathrm{s}$) & \revise{Annual, Monthly} & MERRA-2 (M2T1NXFLX) \\
\texttt{Sr\_b1} & Band 1 (ultra blue, coastal aerosol) surface reflectance & \revise{Annual, Monthly} & Landsat 9 \\
\texttt{Sr\_b2} & Band 2 (blue) surface reflectance	& \revise{Annual, Monthly} & Landsat 9 \\
\texttt{Sr\_b3} & Band 3 (green) surface reflectance & \revise{Annual, Monthly} & Landsat 9 \\
\texttt{Sr\_b4} & Band 4 (red) surface reflectance & \revise{Annual, Monthly} & Landsat 9 \\
\texttt{Sr\_b5} & Band 5 (near infrared) surface reflectance & \revise{Annual, Monthly} & Landsat 9 \\
\texttt{Sr\_b6} & Band 6 (shortwave infrared 1) surface reflectance & \revise{Annual, Monthly} & Landsat 9 \\
\texttt{Sr\_b7} & Band 7 (shortwave infrared 2) surface reflectance & \revise{Annual, Monthly} & Landsat 9 \\
\texttt{Ts} & Surface skin temperature (°C) & \revise{Annual, Monthly} & MERRA-2 (M2T1NXSLV) \\
\texttt{Ulml} & Surface eastward wind ($\mathrm{m}/\mathrm{s}$) & \revise{Annual, Monthly} & MERRA-2 (M2T1NXFLX) \\
\texttt{Vaa} & View Azimuth Angle (°) & \revise{Annual} & HLSL30 (HLS-2) \\
\texttt{Vlml} & Surface northward wind speed ($\mathrm{m}/\mathrm{s}$) & \revise{Annual, Monthly} & MERRA-2 (M2T1NXFLX) \\
\texttt{Vza} & View Zenith Angle (°) & \revise{Annual} & HLSL30 (HLS-2) \\
\texttt{Mean\_2m\_air\_temperature} & Average air temperature at 2m height (°C) & \revise{Annual, Monthly} & ERA5 (Daily Aggregates) \\
\texttt{Mean\_sea\_level\_pressure} & Mean sea level pressure (Pa) & \revise{Annual, Monthly} & ERA5 Daily Aggregates \\
\texttt{Surface\_pressure} & Surface pressure (Pa) & \revise{Annual, Monthly} & ERA5 (Daily Aggregates) \\
\texttt{Total\_precipitation} & Total precipitation (\revise{mm}) & \revise{Annual, Monthly} & ERA5 (Daily Aggregates) \\
\revise{\texttt{NDVI}} & \revise{Normalized Difference Vegetation  Index} & \revise{Annual, Monthly} & \revise{Landsat 9} \\
\revise{\texttt{NDWI}} & \revise{Normalized Difference Water Index} & \revise{Annual, Monthly} & \revise{Landsat 9} \\
\revise{\texttt{NDBI}} & \revise{Normalized Difference Building Index} & \revise{Annual, Monthly} & \revise{Landsat 9} \\
\bottomrule\end{tabular}
}
\label{tab:rswk_rsk}
\end{table}

\begin{table}[ht]
\centering
\caption{World Knowledge Fields and Descriptions for Landmark in the RSWK dataset.}
\resizebox{0.5\textwidth}{!}{
\begin{tabular}{lp{7cm}}
\toprule
\textbf{Field Name} & \textbf{Description} \\
\midrule
\texttt{Name} & The official or commonly known name of the landmark. \\
\texttt{Category} & The type of the landmark. \\
\texttt{Area} & The general geographical region or country where the landmark is located. \\
\texttt{Location} & Latitude and longitude coordinates of the landmark. \\
\texttt{Address} & Full postal or descriptive address of the landmark. \\
\texttt{Physical\_Area} & The spatial footprint or size of the landmark. \\
\texttt{Construction\_Period} & The years or range of years during which the landmark was constructed. \\
\texttt{Historical\_Background} & Historical context, including origin, founding date, and development over time. \\
\texttt{Major\_Events} & Key historical or cultural events that occurred at the landmark. \\
\texttt{Architectural\_Characteristics} & Notable features such as design style, construction materials, and structure. \\
\texttt{Cultural\_Significance} & The cultural, symbolic, or religious importance of the landmark. \\
\texttt{Primary\_Function} & The main use or role of the landmark (e.g., tourist attraction, government building). \\
\texttt{Notable\_Visitors} & Famous individuals or groups who have visited the site. \\
\texttt{Details} & Additional facts or interesting information not captured by other fields. \\
\bottomrule
\end{tabular}}
\label{tab:rswk_wk}
\end{table}

\subsection{Data Statistics} 

\revise{Through the above data construction pipeline, the RSWK dataset successfully collected a total of 14,820 landmark instances from 184 countries, each accompanied by high-resolution satellite imagery, structured remote sensing knowledge, and comprehensive world knowledge descriptions. The imagery is categorized into 16 representative scene classes, including Airport, Amusement Park, Beach, Bridge, Casino, Church, Government Building, Historic Site, Mansion, Museum, Park, Stadium, Theater, Tower, University, and Forest. This categorization captures both urban and natural scenes, enabling the dataset to support a wide range of geospatial vision-language tasks. The remote sensing knowledge module comprises 39 expert-derived variables, as detailed in Table~\ref{tab:rswk_rsk}, including surface reflectance, land surface temperature, albedo, precipitation, and soil moisture. In parallel, each landmark is enriched with 14 structured world knowledge fields, as shown in Table~\ref{tab:rswk_wk}, capturing descriptive and contextual information such as Historical Background, Cultural Significance, Architectural Characteristics, and Primary Function.}

The global spatial distribution of the landmarks is visualized in Fig.2(a), demonstrating wide coverage across all major continents. Notably, the dataset includes landmarks from a diverse range of regions, effectively covering most major countries worldwide. To provide a more detailed view of the dataset composition, we present the distribution of landmarks across the top 50 countries sorted by landmark count in Fig.2(b, left). It can be observed that countries such as the United States, the United Kingdom, China, and Japan contribute the largest number of landmarks. This reflects both the cultural richness and the degree of documentation available for landmarks in these regions. In addition, Fig.2(b, right) shows the frequency distribution of the top 16 landmark categories. The category distribution is relatively balanced, highlighting the diverse types of landmarks captured in the dataset and ensuring a rich set of semantic concepts for downstream tasks. To provide an intuitive example of how the data is structured, we present the Sydney Opera House in Fig.2(c) as a representative landmark. The figure illustrates the three core components of each RSWK entry: the satellite imagery, the domain knowledge, and the structured world knowledge. This tri-modal representation demonstrates the depth and richness of the dataset, supporting a wide range of geospatial understanding and vision-language reasoning tasks.



\begin{table}[h] \color{black}
\centering
\caption{Overview of dataset statistics across task categories in both full and mini benchmark settings.}
\resizebox{0.48\textwidth}{!}{
\begin{tabular}{l c c c c c c c}
\toprule
\multicolumn{2}{l}{\textbf{Task Category}} & \textbf{Total (Full)} & \textbf{Train (Full)} & \textbf{Test (Full)} & \textbf{Total (Mini)} & \textbf{Train (Mini)} & \textbf{Test (Mini)} \\
\midrule
\multicolumn{2}{l}{{Image Captioning}} & {14820}     & 11827 & 2993 &  7408  &5912 & 1496 \\
\multicolumn{2}{l}{{Image Classification}} & {14820} & 11827 & 2993 & 4800 & 3200 & 1600 \\
\multirow{3}{*}{VQA} & {VQA\_C}    & {22604} & 18103 & 4501  & 11301 & 9051 & 2250 \\
& {VQA\_RSK} &  {14820} & 11827 & 2993 &  7408 & 5912 & 1496 \\
& {VQA\_WK} &  {14820} & 11827 & 2993 &  7409 & 5913 & 1496 \\
\bottomrule
\end{tabular}}
\label{tab:benchmark_stats}
\end{table}

{\color{black}
\subsection{Benchmark Setting}

To facilitate systematic evaluation of VLMs in the context of remote sensing and world knowledge understanding, we establish a comprehensive benchmark suite based on the RSWK dataset. This benchmark encompasses three core tasks: image captioning, image classification, and VQA. Specifically, the dataset includes 14,820 image captioning samples, each paired with a descriptive sentence that summarizes the visual content of the satellite imagery along with relevant remote sensing expert knowledge and contextual world knowledge. It also includes 14,820 image classification question-answer pairs, where each question queries the semantic category of the depicted scene, and the answer corresponds to a predefined class label. For the VQA task, we construct a total of 52,244 question-answer pairs aimed at evaluating the model’s multimodal reasoning abilities. These VQA instances are further divided into three types: (1) VQA\_C: 22,604 comprehensive questions that require holistic understanding of both visual content and combined remote sensing and world knowledge; (2)  VQA\_RSW: 14,820 remote sensing knowledge questions that involve interpreting scientific and geophysical attributes; and (3)  VQA\_WK: 14,820 world knowledge questions that focus on encyclopedic or factual information. To support model development and reproducible evaluation, we further split the dataset into 11,827 training samples and 2,993 testing samples. The exact sample counts for each task type are summarized in Table~\ref{tab:benchmark_stats}. Additionally, considering the computational constraints faced by some researchers, we construct a mini-benchmark version comprising approximately 50\% of the full data for each task category, with detailed statistics also reported in Table~\ref{tab:benchmark_stats}.
}

\begin{figure*}
    \centering
    \includegraphics[width=1.0\linewidth]{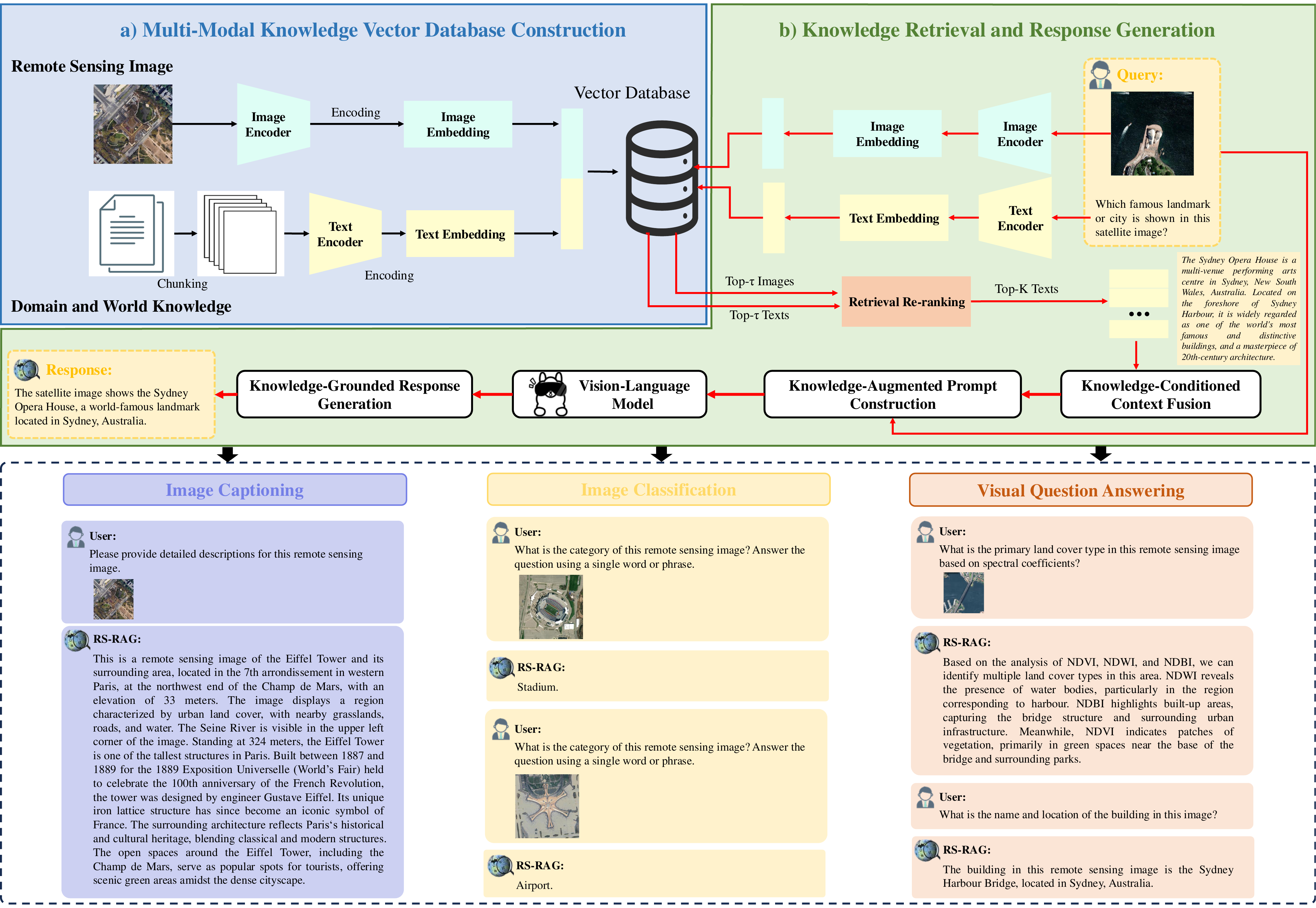}
    \caption{Overview of the proposed Remote Sensing Retrieval-Augmented Generation (RS-RAG) model. It consists of two main processes: (a) The Multi-Modal Knowledge Vector Database Construction module encodes remote sensing imagery and domain/world knowledge into a unified vector space via image and text encoders, enabling efficient cross-modal retrieval. (b) The Knowledge Retrieval and Response Generation module retrieves top-k relevant knowledge based on image and/or textual queries, and re-ranks the results for better relevance. Retrieved knowledge is fused into the prompt through Knowledge-Conditioned Context Fusion, guiding the Vision-Language Model (VLM) to generate Knowledge-Grounded Responses. The RS-RAG model supports multiple downstream tasks such as Image Captioning, Image Classification, and Visual Question Answering, as demonstrated in the bottom section. }
    \label{fig:rsrag}
\end{figure*}

\section{Methodology}
To bridge the semantic gap between remote sensing imagery and comprehensive external knowledge, we propose RS-RAG, a Retrieval-Augmented Generation framework designed to integrate both domain-specific and world knowledge into vision-language reasoning. As illustrated in Fig.~\ref{fig:rsrag}, RS-RAG consists of two main components: the \textit{Multi-Modal Knowledge Vector Database Construction} module, which encodes remote sensing imagery and textual knowledge into a unified embedding space; and the \textit{Knowledge Retrieval and Response Generation} module, which retrieves and fuses the most relevant knowledge to support downstream tasks. By conditioning the vision-language model on retrieved context, RS-RAG enables knowledge-grounded understanding for diverse applications such as image captioning, scene classification, and visual question answering.

\subsection{Problem Formulation}

VLMs are designed to generate natural language outputs conditioned on multimodal inputs, typically a visual observation and a textual prompt. Let \( q_I \) denote the input image (e.g., a remote sensing image), and \( q_T \) the associated textual prompt (e.g., a question or instruction). A conventional VLM seeks to generate a natural language response \( \hat{y} \) by modeling the conditional probability distribution over output space and selecting the most probable response. Formally, this can be expressed as:
\begin{equation}
\hat{y} = \arg\max_{y} P(y \mid q_T, q_I; \theta_{\text{VLM}}),
\label{eq:vlm_standard}
\end{equation}
where \( \theta_{\text{VLM}} \) represents the parameters of the VLM, \( \hat{y} \) can represent an image caption, a classification label, or an answer to a visual question. This closed-form generation framework assumes that all necessary information for reasoning is either visually grounded in the input image \( q_I \) or implicitly encoded within the model parameters \( \theta_{\text{VLM}} \). While this assumption often holds for natural image datasets, it becomes problematic in the domain of remote sensing. Remote sensing images typically capture large-scale scenes, such as entire cities or extensive geographic regions, where accurate interpretation often relies on understanding cultural, historical, or geographical significance. Such knowledge is rarely discernible from visual features and is not explicitly modeled in conventional VLMs.

To address this limitation, we extend the standard VLM formulation by incorporating external, query-relevant knowledge. Specifically, we adopt a \revise{RAG} framework, wherein a set of top-\(k\) relevant knowledge snippets \( \mathbf{R} \) is retrieved from an external corpus based on the multimodal similarity between the input image–text pair \( (q_I, q_T) \) and the knowledge index. The retrieved context \( \mathbf{R} \) is then integrated with the original inputs to guide the response generation process. Formally, the objective becomes:
\begin{equation}
\hat{y} = \arg\max_{y} P(y \mid q_T, q_I, \mathbf{R}; \theta_{\text{VLM}}),
\label{eq:vlm_rag}
\end{equation}
where the additional context \( \mathbf{R} \) allows the model to produce more informative and contextually grounded outputs. This open-book generation paradigm is particularly well-suited for remote sensing applications, where high-level semantic understanding often depends on both domain-specific knowledge (e.g., land use categories) and broader world knowledge (e.g., cultural or geopolitical significance)—information that is typically absent from raw pixel data alone.

\subsection{Multi-Modal Knowledge Vector Database Construction}

To enable retrieval-augmented generation, we construct a Multi-Modal Knowledge Vector Database (MKVD) by encoding the RSWK dataset, which contains high-resolution remote sensing images \(I_i\) and their paired textual descriptions \(T_i\). These data are transformed into dense embeddings using CLIP and stored in a shared semantic space to support efficient and flexible cross-modal retrieval. We adopt CLIP as a unified encoder consisting of a image encoder \(f_{\text{I}}(\cdot)\) and a text encoder \(f_{\text{T}}(\cdot)\), each mapping input into a shared embedding space \(\mathbb{R}^d\). Each image \(I_i\) is encoded into a visual embedding:
\begin{equation}
\mathbf{v}_i = f_{\text{I}}(I_i) \in \mathbb{R}^d.
\end{equation}
In parallel, the corresponding textual document \(T_i\) is segmented into \(m_i\) semantically coherent chunks \(\{T_{i,1}, T_{i,2}, \dots, T_{i,m_i}\}\), and each chunk is encoded using the text encoder:
\begin{equation}
\mathbf{t}_{i,j} = f_{\text{T}}(T_{i,j}) \in \mathbb{R}^d.
\end{equation}

The resulting image embeddings \(\{\mathbf{v}_i\}\) and text embeddings \(\{\mathbf{t}_{i,j}\}\) are indexed in {Qdrant}, a high-performance vector database optimized for approximate nearest neighbor (ANN) search. To organize the data, embeddings are stored in two separate collections: \(\mathcal{D}_{\text{image}}\) for image embeddings and \(\mathcal{D}_{\text{text}}\) for text embeddings. Each image-text pair is linked via a unique identifier \(ID_i\). Specifically, each \(\mathbf{v}_i \in \mathcal{D}_{\text{image}}\) is associated with a set \(\{\mathbf{t}_{i,j}\}_{j=1}^{m_i} \subset \mathcal{D}_{\text{text}}\), which encodes domain-specific and general world knowledge about the same location or object. Metadata such as raw textual descriptions, image paths, and geospatial attributes are stored alongside each entry as payloads. This database structure supports modality-specific and cross-modal retrieval, serving as the foundation for external knowledge integration in the RS-RAG framework.

\subsection{Knowledge Retrieval and Response Generation}

After constructing the Multi-Modal Knowledge Vector Database, we implement a retrieval-augmented generation pipeline that enhances vision-language understanding by incorporating external knowledge retrieved via cross-modal similarity. Given a user query \( q \), composed of an image component \( q_{\text{I}} \) and a textual component \( q_{\text{T}} \), we first encode each input into dense embeddings:
\begin{equation}
\mathbf{v}_{\text{T}} = f_{\text{T}}(q_{\text{T}}), \quad \mathbf{v}_{\text{I}} = f_{\text{I}}(q_{\text{I}}),
\end{equation}
where \( f_{\text{T}}(\cdot) \) and \( f_{\text{I}}(\cdot) \) denote the CLIP-based text and image encoders, respectively. To retrieve semantically relevant knowledge, we perform similarity search in both the text and image embedding spaces, retrieving the top-\( \tau \) candidates from each modality:
\begin{equation}
\mathcal{R}_{\text{T}}^\tau = \text{Top}_\tau(\mathbf{v}_{\text{T}}, \mathcal{D}_{\text{text}}), \quad \mathcal{R}_{\text{I}}^\tau = \text{Top}_\tau(\mathbf{v}_{\text{I}}, \mathcal{D}_{\text{image}}),
\end{equation}
where \( \mathcal{D}_{\text{text}} \) and \( \mathcal{D}_{\text{image}} \) represent the text and image embedding collections in the vector database. While the initial retrieval from each modality yields candidates based on unimodal similarity, these results may contain semantically redundant, irrelevant, or inconsistent entries due to the disjoint nature of visual and textual embedding spaces. To address this issue, we introduce a \textit{retrieval re-ranking} step that jointly considers both modalities. Specifically, we first merge the retrieved sets from each modality, $\mathcal{R}_{\text{fused}} = \mathcal{R}_{\text{T}}^\tau \cup \mathcal{R}_{\text{I}}^\tau$. Each candidate is then assigned a fused similarity score via weighted combination:
\begin{equation} \label{eq:score}
\text{score}(r_i) = (1 - \alpha) \cdot s_{\text{T}}(r_i) + \alpha \cdot s_{\text{I}}(r_i),
\end{equation}
where $s_{\text{text}}(r_i)$ and $s_{\text{image}}(r_i)$ are the cosine similarities between the query and the candidate in the respective embedding spaces. The weighting parameter $\alpha \in [0, 1]$ controls the relative influence of each modality.Based on these fused scores, we select the top-\( K \) most relevant candidates:
\begin{equation}
\{k_1, \dots, k_K\} = \text{Top}_K\left( \{ \text{score}(r_i) \mid r_i \in \mathcal{R}_{\text{fused}} \} \right).
\end{equation}
To enhance semantic coherence and eliminate redundancy among the retrieved segments, we apply a \textit{Knowledge-Conditioned Context Fusion} module that consolidates them into a single, contextually grounded representation. Specifically, a frozen large language model \( \mathcal{L}_{\text{fuse}} \)  is employed to synthesize the  knowledge-conditioned context \( \mathbf{R} \) from the top-ranked knowledge snippets:
\begin{equation}
\mathbf{R} = \mathcal{L}_{\text{fuse}}(\{k_1, \dots, k_K\}).
\end{equation}
This fusion step consolidates salient content from the retrieved segments into a compact, context-aware representation, thereby facilitating structured prompt construction. Given the original user query  \( q_T \)   and the fused knowledge context \( \mathbf{R} \), we construct a retrieval-augmented prompt \( P_q \)  via \textit{Knowledge-Augmented Prompt Construction} as follows:
\begin{equation}
P_q = \text{Concat}\left[\phi,\; q_T,\; \psi,\; \mathbf{R} \right]
\end{equation}
where \( \phi \) is a task-specific instruction token (e.g., ``Answer the following question based on the retrieved knowledge:''), \( \psi \) is a knowledge header (e.g., ``Retrieved context:''), and \( \mathbf{R} \) is the fused knowledge used to support reasoning. Finally, the composed prompt \( P_q \), along with the visual input \( q_{\text{I}} \), is provided to the VLM, which performs Knowledge-Grounded Response Generation via joint multimodal reasoning and generates the final output:
\begin{equation}
\hat{y} = \text{Generate}_{\theta_{\text{VLM}}}(y \mid \text{Image}=q_{\text{I}},\; \text{Prompt}=P_q).
\end{equation}
By leveraging retrieval from a multi-modal knowledge base, this framework empowers the RSVLMs to go beyond purely visual grounding by integrating both domain-specific knowledge and broader world knowledge, including cultural, historical, and geopolitical context. This enriched understanding enables the model to generate more accurate, context-aware, and semantically comprehensive outputs. In doing so, our RS-RAG framework effectively bridges the gap between remote sensing imagery and external knowledge sources, establishing a retrieval-augmented generation paradigm tailored to the unique demands of the remote sensing domain.

\begin{figure*} \color{black}
    \centering
    \includegraphics[width=1.0\linewidth]{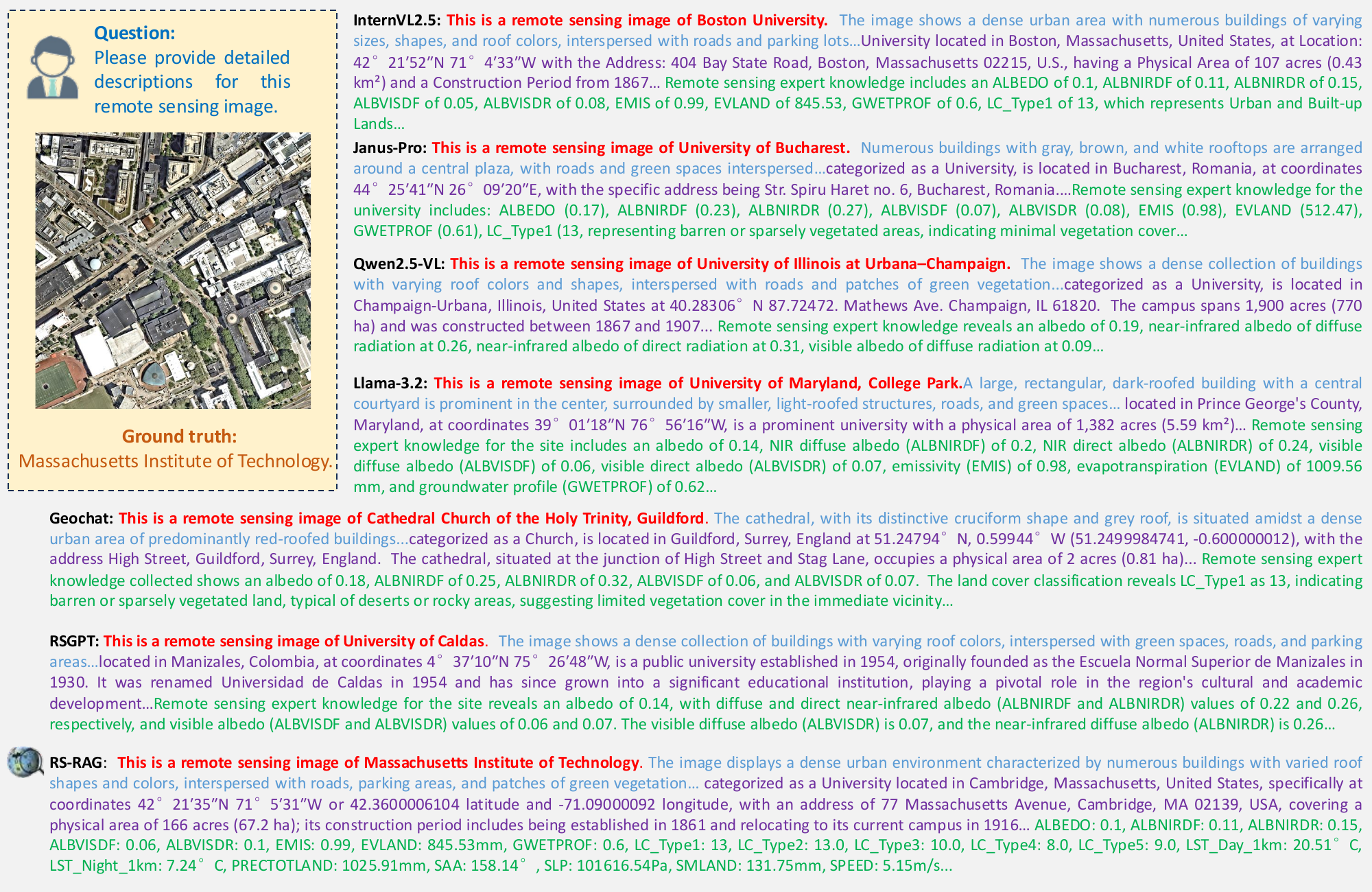}
    \caption{Qualitative results of image captioning on remote sensing imagery of the Massachusetts Institute of Technology (MIT). Text in red indicates the identified landmark name, blue highlights visual content directly observable from the image, purple denotes retrieved world knowledge, such as historical, geographic, or cultural facts; and green denotes remote sensing knowledge, including spectral indices, land cover, and ALBEDO values.}
    \label{fig:res_cap}
\end{figure*}

\begin{table*}[t] \color{black}
\centering
\small
\caption{Performance comparison between baseline models and our proposed RS-RAG variants on the image captioning task using the RSWK-Mini dataset.}
\begin{tabular}{lccccccc}
\toprule
\textbf{Model} & \textbf{BLEU-1} & \textbf{BLEU-2} & \textbf{BLEU-3} & \textbf{BLEU-4} & \textbf{METEOR} & \textbf{ROUGE-L} & \textbf{CIDEr} \\
\midrule
InternVL2.5~\cite{chen2024internvl} & 0.368 & 0.221 & 0.154 & 0.113 & 0.312 & 0.325 & 0.014 \\
Janus-Pro~\cite{chen2025janus} & 0.370 & 0.221 & 0.154 & 0.112 & 0.312 & 0.324 & 0.013 \\
Qwen2.5-VL~\cite{qwen2.5-VL} & 0.385 & 0.237 & 0.169 & 0.125 & 0.331 & 0.344 & 0.018 \\
Llama3.2-Vision~\cite{meta2024llama} & 0.379 & 0.232 & 0.164 & 0.122 & 0.321 & 0.338 & 0.016 \\
RSGPT~\cite{hu2025rsgpt} & 0.192 & 0.097 & 0.061 & 0.041 & 0.144 & 0.154 & 0.002 \\
Geochat~\cite{kuckreja2023geochat} & 0.347 & 0.201 & 0.139 & 0.101 & 0.291 & 0.297 & 0.006 \\
\midrule
RS-RAG & \textbf{0.490}& \textbf{0.366} & \textbf{0.300} & \textbf{0.252} & \textbf{0.447} & \textbf{0.476} & \textbf{0.145} \\
\bottomrule
\end{tabular}
\label{tab:res_cap}
\end{table*}

\begin{table*}[h] \color{black}
\centering
\small
\caption{Performance comparison between baseline models and our proposed RS-RAG variants on the image classification task using the RSWK-Mini dataset.}
\resizebox{\textwidth}{!}{
\begin{tabular}{lccccccccccccccccc}
\toprule
\textbf{Model} & \textbf{Overall} & \textbf{Airport} & \textbf{Amusement} & \textbf{Beach} & \textbf{Bridge} & \textbf{Casino} & \textbf{Church} & \textbf{Gov. Bldg.} & \textbf{Historic} & \textbf{Mansion} & \textbf{Museum} & \textbf{Park} & \textbf{Stadium} & \textbf{Theater} & \textbf{Tower} & \textbf{University} &\textbf{Forest} \\
 & \textbf{Accuracy} & & \textbf{Park}  & & & & & & \textbf{Site} & & & & & & & \\
\midrule
InternVL2.5~\cite{chen2024internvl} & 0.57 & 0.91 & 0.59 & 0.73 & 0.76 & 0.82 & 0.35 & 0.42 & 0.27 & 0.60 & 0.16 & 0.21 & 0.90 & 0.45 & 0.41 & 0.72 & 0.86 \\
Janus-Pro~\cite{chen2025janus} & 0.58 & 0.95 & 0.67 & 0.62 & 0.84 & 0.76 & 0.40 & 0.37 & 0.17 & 0.64 & 0.30 & 0.20 & 0.90 & 0.46 & 0.49 & 0.64 & 0.87 \\
Qwen2.5-VL~\cite{qwen2.5-VL} & 0.61 & 0.90 & 0.66 & 0.69 & 0.78 & 0.77 & 0.51 & 0.39 & 0.23 & 0.70 & 0.29 & 0.32 & 0.88 & 0.52 & \textbf{0.59} & 0.69 & 0.86 \\
Llama3.2-Vision~\cite{meta2024llama} & 0.59 & 0.93 & 0.62 & 0.66 & 0.77 & 0.72 & 0.43 & 0.39 & 0.13 & 0.75 & 0.41 & 0.28 & 0.90 & 0.48 & 0.57 & 0.55 & \textbf{0.90} \\
Rsgpt~\cite{hu2025rsgpt} & 0.46 & 0.91 & 0.70 & 0.79 & 0.75 & 0.37 & 0.22 & 0.18 & 0.31 & 0.50 & 0.06 & 0.49 & 0.93 & 0.06 & 0.11 & 0.66 & 0.39 \\
Geochat~\cite{geochat} & 0.60 & 0.94 & 0.55 & 0.75 & 0.79 & 0.79 & 0.54 & 0.34 & 0.11 & 0.65 & 0.27 & 0.39 & 0.88 & \textbf{0.6} & 0.57 & 0.64 & 0.84 \\
\midrule
RS-RAG & \textbf{0.79} & \textbf{0.98} & \textbf{0.76} & \textbf{0.77} & \textbf{0.92} & \textbf{0.91} & \textbf{0.81} & \textbf{0.98} & \textbf{0.75} & \textbf{0.92} & \textbf{0.72} & \textbf{0.80} & \textbf{0.99} & 0.23 & 0.52 & \textbf{0.74} & 0.89 \\
\bottomrule
\end{tabular}}
\label{tab:res_cls}
\end{table*}

\begin{figure}[h] \color{black}
    \centering
    \includegraphics[width=1.0\linewidth]{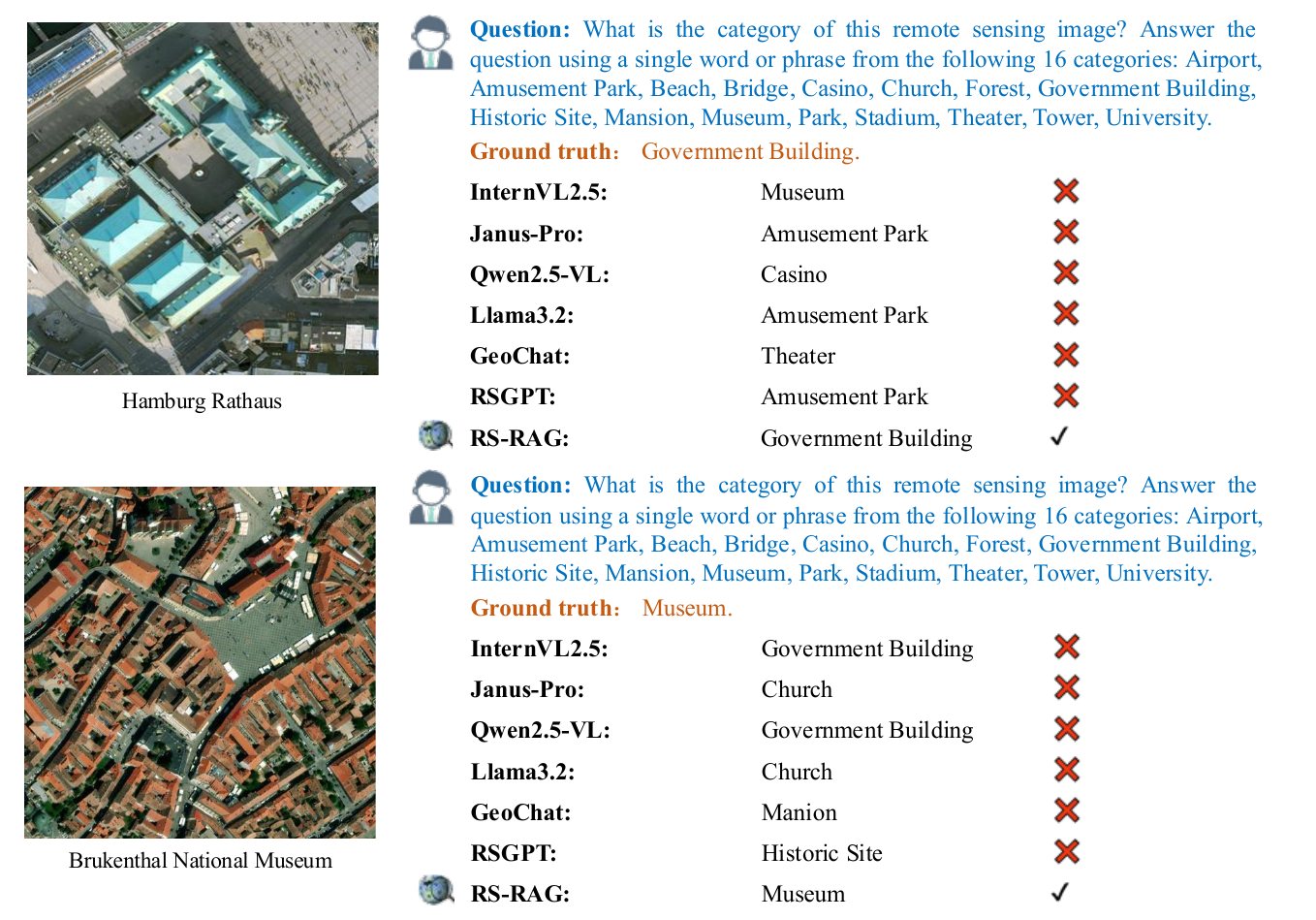}
    \caption{ Qualitative results of baseline models and our RS-RAG model on the image classification task using the RSWK dataset.
}
    \label{fig:res_cls}
\end{figure}

\section{Experiments}


\subsection{Experimental Setup}

\revise{All input images are resized to 512 by 512 pixels before being fed into the models. We evaluate several state-of-the-art vision-language models as baselines, including {InternVL2.5-Instruct-8B~\cite{chen2024internvl}}, {Janus-Pro-7B~\cite{chen2025janus}}, {Qwen-2.5-VL-7B-Instruct~\cite{qwen2.5-VL}}, {LLaMA-3.2-Vision-11B-Instruct~\cite{meta2024llama}}, RSGPT~\cite{hu2025rsgpt}, and Geochat~\cite{geochat}. Our proposed RS-RAG model is built upon Qwen-2.5-VL-7B. To better adapt these models to the remote sensing domain, we perform instruction tuning on the RSWK dataset using Low-Rank Adaptation (LoRA)~\cite{hu2022lora}.  Fine-tuning is conducted for 3 epochs with a batch size of 1, using the Adamw optimizer and an initial learning rate of \(1 \times 10^{-4}\). All experiments are performed using 3 NVIDIA RTX A6000 GPUs, each with 48 GB of memory.}

To comprehensively evaluate model performance, we adopt a set of standard metrics tailored to each task. For image captioning and visual question answering, we report BLEU-1, BLEU-2, BLEU-3, BLEU-4, METEOR, ROUGE-L, and CIDEr to measure the fluency, relevance, and informativeness of generated text. For the classification task, we evaluate both overall accuracy and per-class accuracy to reflect model performance across diverse scene categories.

\subsection{Results on Image Captioning Task}

\revise{After instruction tuning all VLMs, we first evaluate their performance on the image captioning task. Table~\ref{tab:res_cap} presents a comparative analysis of baseline VLMs and our proposed RS-RAG model on the RSWK-Mini dataset. Across all seven evaluation metrics, BLEU-1 to BLEU-4, METEOR, ROUGE-L, and CIDEr, RS-RAG consistently outperforms all baselines by a substantial margin. While Qwen2.5-VL surpasses other baseline models, RS-RAG achieves a BLEU-1 score of 0.490 and a CIDEr score of 0.145, representing improvements of 10.5 and 12.7 points respectively over Qwen2.5-VL. These results highlight the benefit of incorporating structured external knowledge into VLMs, which significantly enhances their capacity to generate accurate, contextually rich, and semantically meaningful captions in the remote sensing domain. The superior performance of RS-RAG underscores the effectiveness of our retrieval-augmented design for complex, knowledge-intensive multimodal reasoning tasks.}

\revise{To better understand the effectiveness of our RS-RAG framework, we conduct a qualitative comparison of image captioning outputs generated by baseline models and our RS-RAG model, as shown in Fig.\ref{fig:res_cap}. Specifically, Fig.\ref{fig:res_cap} presents a remote sensing image corresponding to the Massachusetts Institute of Technology (MIT). Among all evaluated models, RS-RAG is the only one that correctly identifies the landmark and accurately integrates both factual and scientific domain knowledge into the caption. The output from RS-RAG includes detailed metadata such as the official address, spatial coverage, historical construction timeline, and multiple spectral and physical indicators. This demonstrates RS-RAG’s ability to effectively synthesize geographic information with remote sensing expertise. In contrast, the baseline models exhibit limitations in both semantic accuracy and factual grounding. For instance, InternVL2.5 misidentifies the site as Boston University, GeoChat assigns the location to Cathedral Church, and RSGPT attributes it to the University of Caldas in Colombia. Some baseline models do retrieve domain-relevant attributes, yet the outputs often lack factual coherence due to incorrect or inconsistent landmark recognition. These findings highlight the advantages of RS-RAG in connecting visual content with structured multimodal knowledge representations. The model demonstrates strong capability in supporting knowledge-intensive remote sensing tasks, particularly those requiring both precise semantic understanding and accurate factual alignment.}

\subsection{Results on Image Classification Task}

\revise{We further evaluate the capabilities of vision-language models on the image classification task using the RSWK-Mini dataset. As shown in Table~\ref{tab:res_cls}, RS-RAG significantly outperforms all baseline models in terms of overall accuracy as well as performance across most individual categories. Specifically, RS-RAG achieves an overall accuracy of 0.79, representing a substantial improvement over the strongest baseline, Qwen2.5-VL, which achieves 0.61. In terms of category-specific performance, RS-RAG consistently delivers top results in 13 out of 16 categories, with particularly high accuracies for Airport (0.98), Stadium (0.99), Mansion (0.92), and Park (0.80). For commonly encountered scenes in remote sensing, such as Forest, although RS-RAG does not achieve the highest score, its accuracy of 0.89 still demonstrates strong capability in handling typical geospatial scenarios. These results underscore the effectiveness of our retrieval-augmented design in enhancing the scene recognition capacity of VLMs, particularly in remote sensing contexts that demand fine-grained semantic understanding. To further evaluate RS-RAG’s capability in scene classification, we conduct a qualitative comparison on two challenging urban examples, as shown in Fig.~\ref{fig:res_cls}. The task requires selecting the correct category from 16 visually and semantically similar classes. In the first case, RS-RAG correctly classifies the Hamburg Rathaus as a Government Building, while all baseline models fail, offering incorrect labels like “Museum” or “Amusement Park.” In the second case, only RS-RAG accurately identifies the Brukenthal National Museum, whereas other models misclassify it as a Church or Historic Site. These results demonstrate RS-RAG’s superior ability to align visual content with structured knowledge, enabling robust performance in semantically ambiguous remote sensing scenes. }

\begin{figure}[ht] \color{black}
    \centering
    \includegraphics[width=1.0\linewidth]{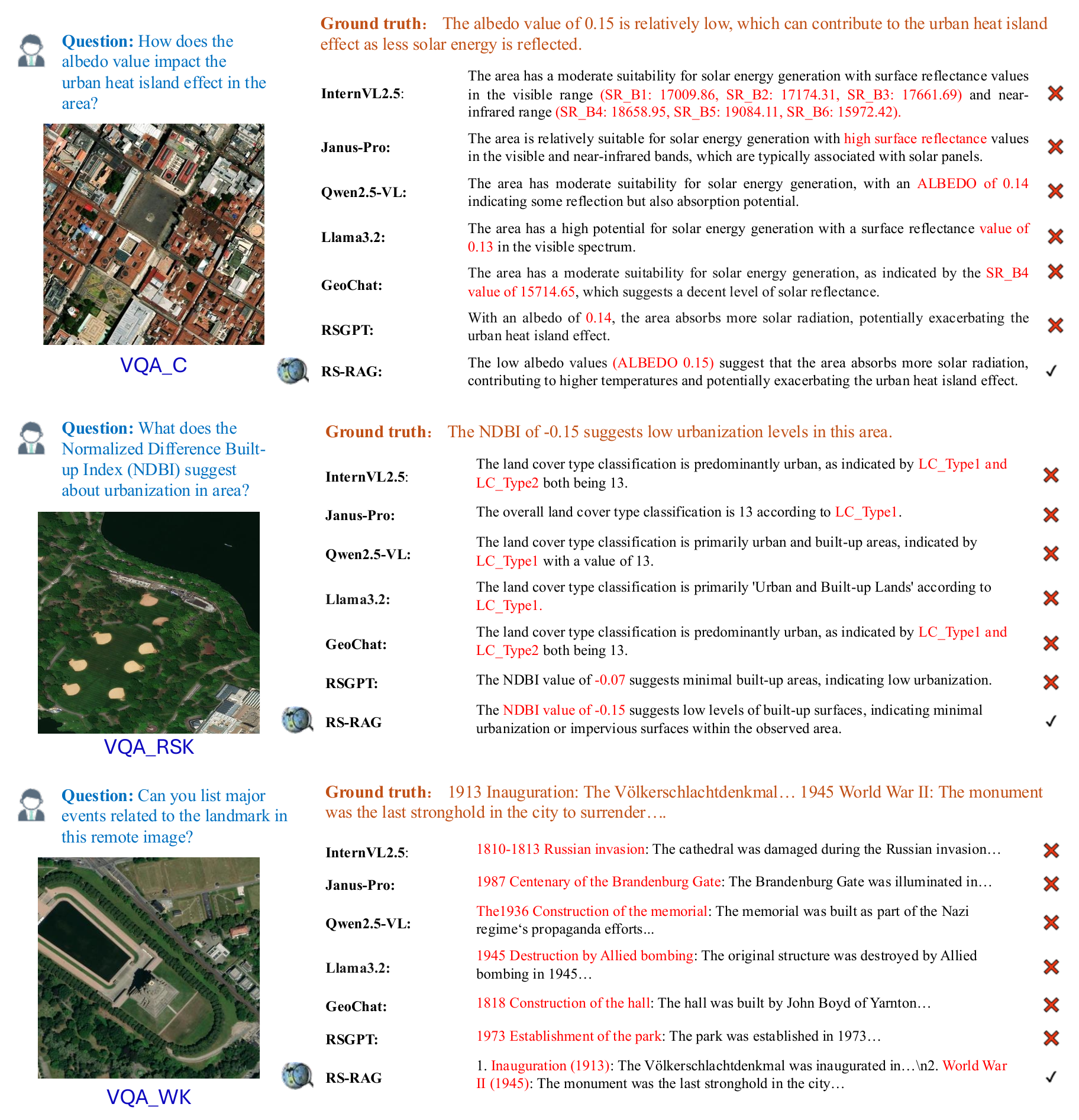}
    \caption{Qualitative results of baseline models and our RS-RAG model on the VQA task using the RSWK dataset.}
    \label{fig:res_vqa}
\end{figure}

\begin{table*}[h] \color{black}
\centering
\small
\caption{Performance comparison between baseline models and our proposed RS-RAG variants on the image VQA task using the RSWK-Mini dataset}
\begin{tabular}{lccccccc}
\toprule
\textbf{Model} & \textbf{BLEU-1} & \textbf{BLEU-2} & \textbf{BLEU-3} & \textbf{BLEU-4} & \textbf{METEOR} & \textbf{ROUGE-L} & \textbf{CIDEr} \\
\midrule
InternVL2.5~\cite{chen2024internvl} & 0.095 & 0.038 & 0.024 & 0.018 & 0.078 & 0.182 & 0.134 \\
Janus-Pro~\cite{chen2025janus} & 0.094 & 0.038 & 0.024 & 0.017 & 0.076 & 0.180 & 0.131 \\
Qwen2.5-VL~\cite{qwen2.5-VL} & 0.102 & 0.041 & 0.027 & 0.019 & 0.083 & 0.187 & 0.158 \\
Llama3.2-Vision~\cite{meta2024llama} & 0.102 & 0.042 & 0.027 & 0.019 & 0.082 & 0.187 & 0.153 \\
RSGPT~\cite{hu2025rsgpt} & 0.209 & 0.132 & 0.097 & 0.072 & 0.201 & 0.347 & 0.569 \\
Geochat~\cite{kuckreja2023geochat} & 0.207 & 0.129 & 0.093 & 0.068 & 0.193 & 0.353 & 0.531 \\
\midrule
RS-RAG & \textbf{0.254} & \textbf{0.149} & \textbf{0.108} & \textbf{0.080} & \textbf{0.239} & \textbf{0.375} & \textbf{0.685} \\
\bottomrule
\end{tabular}
\label{tab:res_vqa}
\end{table*}

\subsection{Results on Visual Question Answering Task}

\revise{To assess the models’ ability to handle knowledge-intensive multimodal reasoning, we conduct evaluations on the VQA task using the RSWK-Mini dataset. As shown in Table~\ref{tab:res_vqa}, RS-RAG delivers superior performance across all evaluation metrics. It achieves a BLEU-1 score of 0.254 and a CIDEr score of 0.685, significantly surpassing the strongest baseline, RSGPT, by 4.5 and 11.6 points, respectively. Although RSGPT exhibits relatively strong performance, RS-RAG consistently outperforms it across all metrics, especially in higher-order BLEU scores and METEOR, which reflect deeper semantic understanding and language fluency. These results underscore the effectiveness of our retrieval-augmented strategy in integrating both remote sensing knowledge and structured world knowledge into the vision-language reasoning process. The consistent improvements across BLEU, METEOR, ROUGE-L, and CIDEr indicate that RS-RAG not only retrieves relevant content but also generates contextually coherent and semantically meaningful answers. }

\revise{To further illustrate the model’s effectiveness, we provide qualitative examples of the three types of VQA tasks in Fig.~\ref{fig:res_vqa}. RS-RAG consistently generates answers that are not only factually accurate but also grounded in structured remote sensing and world knowledge. In the first example, which corresponds to comprehensive reasoning, RS-RAG correctly interprets the impact of low albedo on the urban heat island effect by incorporating retrieved scientific knowledge, whereas other models either hallucinate surface reflectance values or fail to link them to urban climate phenomena. In the second example, which tests remote sensing knowledge, RS-RAG accurately interprets the meaning of a low NDBI value as an indicator of low urbanization, whereas most baselines confuse land cover classification with index interpretation. In the third example, which focuses on world knowledge, RS-RAG identifies both the historical event and its relevance to the landmark, while other models misattribute unrelated events or fail to provide temporally aligned responses. These results highlight the strength of RS-RAG in connecting visual content with precise, knowledge-grounded answers across different question types, demonstrating its robustness in addressing complex multimodal reasoning within the remote sensing domain.}

\subsection{Discussion}

\textbf{1) Effect of the number of retrieved candidates.}
\revise{To understand the influence of retrieval granularity on downstream performance, we conduct an ablation study by varying the number of top-$k$ retrieved candidates in RS-RAG across three tasks: image captioning, image classification, and VQA, as presented in Table~\ref{tab:ablation_topk}. As shown in Table~\ref{tab:ablation_topk} (a) for image captioning, setting $k$ to 1 already yields the best performance across most evaluation metrics. Specifically, RS-RAG achieves a BLEU-1 score of 0.490, a METEOR score of 0.447, and a ROUGE-L score of 0.476. When $k$ increases to 3 and 5, performance in BLEU and METEOR declines slightly, while CIDEr improves marginally, suggesting that incorporating additional retrieved knowledge may introduce noise while marginally enhancing informativeness. For image classification, shown in Table~\ref{tab:ablation_topk} (b), a consistent degradation in performance is observed as $k$ increases. The best overall accuracy and CIDEr score are obtained when $k$ is 1, indicating that precise and highly relevant knowledge is essential for accurate scene categorization, whereas retrieving more candidates may introduce irrelevant or distracting context. In contrast, VQA results presented in Table~\ref{tab:ablation_topk} (c) show a clear benefit from broader retrieval. Performance steadily improves as $k$ increases, with the highest scores observed when $k$ is 5. Under this setting, RS-RAG achieves a BLEU-1 score of 0.254, a METEOR score of 0.239, and a CIDEr score of 0.685. This suggests that answering complex and knowledge-intensive questions benefits from access to a wider and more diverse set of evidence. Overall, these results highlight the importance of task-specific retrieval strategies. While minimal retrieval is sufficient for captioning and classification, broader retrieval is critical for effective multimodal reasoning in VQA. The consistent performance of RS-RAG across varying retrieval settings underscores its robustness and adaptability in handling diverse remote sensing tasks with structured knowledge.}

\begin{table}[h] \color{black}
\centering
\caption{Ablation study on the number of top-$k$ retrieved candidates for RS-RAG across three tasks: image captioning, classification, and VQA.}
\resizebox{0.48\textwidth}{!}{
\begin{tabular}{lccccccc}
\toprule
\multicolumn{8}{c}{\textbf{(a) Image Captioning}} \\
\midrule
\textbf{Top-$k$} & BLEU-1 & BLEU-2 & BLEU-3 & BLEU-4 & METEOR & ROUGE-L & CIDEr \\
\textbf{$k=1$} & \textbf{0.490} & \textbf{0.366} & \textbf{0.300} & \textbf{0.252} & \textbf{0.447} & \textbf{0.476} & \textbf{0.145} \\
$k=3$ & 0.483 & 0.358 & 0.293 & 0.245 & 0.439 & 0.472 & 0.147 \\
$k=5$ & 0.485 & 0.362 & 0.297 & 0.250 & 0.441 & 0.474 & 0.173 \\
\midrule
\multicolumn{8}{c}{\textbf{(b) Image Classification}} \\
\midrule
\textbf{Top-$k$} & BLEU-1 & BLEU-2 & BLEU-3 & BLEU-4 & METEOR & ROUGE-L & CIDEr \\
\textbf{$k=1$} & \textbf{0.794} & \textbf{0.358} & \textbf{0.213} & \textbf{0.163} & \textbf{0.466} & \textbf{0.795} & \textbf{2.376} \\
$k=3$ & 0.676 & 0.289 & 0.175 & 0.136 & 0.387 & 0.676 & 1.966 \\
$k=5$ & 0.654 & 0.272 & 0.167 & 0.130 & 0.369 & 0.656 & 1.875 \\
\midrule
\multicolumn{8}{c}{\textbf{(c) Image VQA}} \\
\midrule
\textbf{Top-$k$} & BLEU-1 & BLEU-2 & BLEU-3 & BLEU-4 & METEOR & ROUGE-L & CIDEr \\
$k=1$ & 0.242 & 0.136 & 0.097 & 0.071 & 0.220 & 0.359 & 0.594 \\
$k=3$ & 0.246 & 0.140 & 0.100 & 0.074 & 0.228 & 0.365 & 0.617 \\
\textbf{$k=5$} & \textbf{0.254} & \textbf{0.149} & \textbf{0.108} & \textbf{0.080} & \textbf{0.239} & \textbf{0.375} & \textbf{0.685} \\
\bottomrule
\end{tabular}}
\label{tab:ablation_topk}
\end{table}

\textbf{2) Effect of fusion weight $\alpha$.}
\revise{We further investigate the impact of the modality fusion weight $\alpha$ on downstream performance by varying its value across three tasks: image captioning, image classification, and VQA. As summarized in Table~\ref{tab:ablation_alpha}, this coefficient controls the balance between visual and textual similarity scores in the re-ranking stage of retrieval. In the image captioning task, performance improves consistently as $\alpha$ increases. The highest scores are obtained when $\alpha$ is set to 0.9, yielding a BLEU-1 score of 0.490, a METEOR score of 0.447, and a CIDEr score of 0.145. These results suggest that visual similarity plays a dominant role in generating accurate and contextually appropriate captions for remote sensing imagery. Interestingly, the image classification task exhibits a different pattern. The best performance is achieved at $\alpha$ set to 0.5, where RS-RAG obtains a BLEU-1 score of 0.748, a METEOR score of 0.433, and a CIDEr score of 2.376. This indicates that a more balanced integration of visual and textual modalities is critical for effective scene categorization, possibly due to the complementary nature of semantic and spatial cues. In contrast, the VQA task again favors higher $\alpha$ values. When $\alpha$ is set to 0.9, the model achieves its best performance, with BLEU-1 of 0.246, METEOR of 0.228, and CIDEr of 0.617. These results highlight the importance of visual evidence in answering spatially grounded or visually intensive questions. Overall, these findings underscore the necessity of task-specific fusion strategies. While image captioning and VQA benefit from emphasizing visual similarity, image classification requires a more nuanced balance between modalities. This flexibility further demonstrates the robustness and adaptability of the RS-RAG framework in knowledge-driven remote sensing scenarios.}

\begin{table}[h] \color{black}
\centering
\caption{Ablation study on the fusion weight $\alpha$ in RS-RAG across three tasks: image captioning, classification, and VQA.}
\resizebox{0.48\textwidth}{!}{
\begin{tabular}{lccccccc}
\toprule
\multicolumn{8}{c}{\textbf{(a) Image Captioning}} \\
\midrule
\textbf{$\alpha$} & BLEU-1 & BLEU-2 & BLEU-3 & BLEU-4 & METEOR & ROUGE-L & CIDEr \\
0.3 & 0.364 & 0.218 & 0.153 & 0.113 & 0.311 & 0.317 & 0.024 \\
0.5 & 0.399 & 0.256 & 0.190 & 0.147 & 0.347 & 0.355 & 0.066 \\
0.7 & 0.477 & 0.347 & 0.280 & 0.231 & 0.430 & 0.453 & 0.166 \\
\textbf{0.9} & \textbf{0.490} & \textbf{0.366} & \textbf{0.300} & \textbf{0.252} & \textbf{0.447} & \textbf{0.476} & \textbf{0.145} \\
\midrule
\multicolumn{8}{c}{\textbf{(b) Image Classification}} \\
\midrule
\textbf{$\alpha$} & BLEU-1 & BLEU-2 & BLEU-3 & BLEU-4 & METEOR & ROUGE-L & CIDEr \\
0.3 & 0.669 & 0.302 & 0.179 & 0.137 & 0.393 & 0.669 & 2.004 \\
\textbf{0.5} & \textbf{0.748} & \textbf{0.329} & \textbf{0.197} & \textbf{0.152} & \textbf{0.433} & \textbf{0.748} & \textbf{2.208} \\
0.7 & 0.711 & 0.309 & 0.186 & 0.144 & 0.410 & 0.711 & 2.087 \\
0.9 & 0.676 & 0.289 & 0.175 & 0.136 & 0.387 & 0.676 & 1.966 \\
\midrule
\multicolumn{8}{c}{\textbf{(c) Image VQA}} \\
\midrule
\textbf{$\alpha$} & BLEU-1 & BLEU-2 & BLEU-3 & BLEU-4 & METEOR & ROUGE-L & CIDEr \\
0.3 & 0.213 & 0.122 & 0.088 & 0.064 & 0.209 & 0.342 & 0.507 \\
0.5 & 0.212 & 0.121 & 0.087 & 0.064 & 0.208 & 0.339 & 0.506 \\
0.7 & 0.211 & 0.121 & 0.087 & 0.064 & 0.208 & 0.338 & 0.503 \\
\textbf{0.9} & \textbf{0.246} & \textbf{0.140} & \textbf{0.100} & \textbf{0.074} & \textbf{0.228} & \textbf{0.365} & \textbf{0.617} \\
\bottomrule
\end{tabular}}
\label{tab:ablation_alpha}
\end{table}

\begin{table}[h] \color{black}
\centering
\small
\caption{Performance comparison between baseline models and our proposed RS-RAG variants on the image captioning task using the ChatEarthNet dataset.}
\resizebox{0.48\textwidth}{!}{
\begin{tabular}{lccccccc}
\toprule
\textbf{Model} & \textbf{BLEU-1} & \textbf{BLEU-2} & \textbf{BLEU-3} & \textbf{BLEU-4} & \textbf{METEOR} & \textbf{ROUGE-L} & \textbf{CIDEr} \\
\midrule
InternVL2.5~\cite{chen2024internvl} & 0.369 & 0.213 & 0.134 & 0.082 & 0.302 & 0.300 & 0.076 \\
Janus-Pro~\cite{chen2025janus} & 0.366 & 0.210 & 0.131 & 0.080 & 0.306 & 0.295 & 0.070 \\
Qwen2.5-VL~\cite{qwen2.5-VL} & 0.370 & 0.209 & 0.126 & 0.074 & 0.308 & 0.284 & 0.079 \\
Llama3.2-Vision~\cite{meta2024llama} & 0.370 & 0.214 & 0.136 & 0.085 & 0.306 & 0.304 & 0.077 \\
RSGPT~\cite{hu2025rsgpt} & 0.370 & 0.217 & 0.136 & 0.084 & 0.295 & 0.299 & 0.087 \\
Geochat~\cite{kuckreja2023geochat} & 0.335 & 0.201 & 0.128 & 0.078 & 0.301 & 0.284 & 0.016 \\
\midrule
RS-RAG & \textbf{0.388}& \textbf{0.243} & \textbf{0.153} & \textbf{0.098} & \textbf{0.351} & \textbf{0.356} & \textbf{0.171} \\
\bottomrule
\end{tabular}}
\label{tab:chatearth_cap}
\end{table}

\revise{
\textbf{3) Extension of RS-RAG on ChatEarthNet Dataset.}
To further validate the generalization capability of RS-RAG beyond the RSWK dataset, we conduct additional experiments on the ChatEarthNet dataset~\cite{yuan2024chatearthnet}, a recently proposed large-scale image–text dataset designed for training large vision–language models. As shown in Table~\ref{tab:chatearth_cap}, RS-RAG consistently outperforms all baseline models across all seven evaluation metrics, including BLEU-1 to BLEU-4, METEOR, ROUGE-L, and CIDEr. Specifically, RS-RAG achieves a BLEU-1 score of 0.388 and a CIDEr score of 0.171, marking respective gains of 1.8 and 8.4 points over the strongest baseline, RSGPT. These improvements underscore the robustness of RS-RAG in adapting to new remote sensing datasets and generating semantically rich, knowledge-grounded captions. The consistent performance advantage across metrics further confirms the effectiveness of our retrieval-augmented design in enhancing vision–language understanding, even in domains beyond the original dataset.
}

\rrevise{
\textbf{4) Generalization in Visually Similar Scenes.}
To further evaluate the {generalization capability} of RS-RAG beyond landmark recognition, we conducted additional experiments on {common scenes} that exhibit high visual similarity. Specifically, we collected {1,000 samples each for grassland and farmland} from diverse global regions and compared RS-RAG with the {state-of-the-art VLM Qwen2.5-VL}. These two categories share highly similar visual patterns in satellite imagery, making them challenging to distinguish based solely on visual cues. As shown in Table~\ref{tab:grassland_farmland}, RS-RAG consistently outperforms the baseline across both categories, demonstrating its superior reasoning capability when visual information is insufficient.}

\begin{table}[ht] \color{black}
\centering
\caption{Comparison of classification accuracy on visually similar categories: Grassland and Farmland.}
\label{tab:grassland_farmland}
\begin{tabular}{lccc}
\toprule
{Category} & {Qwen2.5-VL} & {RS-RAG} & {Improvement} \\
\midrule
Grassland & 0.834 & \textbf{0.904} & +7.0\% \\
Farmland  & 0.955 & \textbf{0.990} & +3.5\% \\
\bottomrule
\end{tabular}
\end{table}

\begin{figure}[ht] \color{black}
    \centering
    \includegraphics[width=1.0\linewidth]{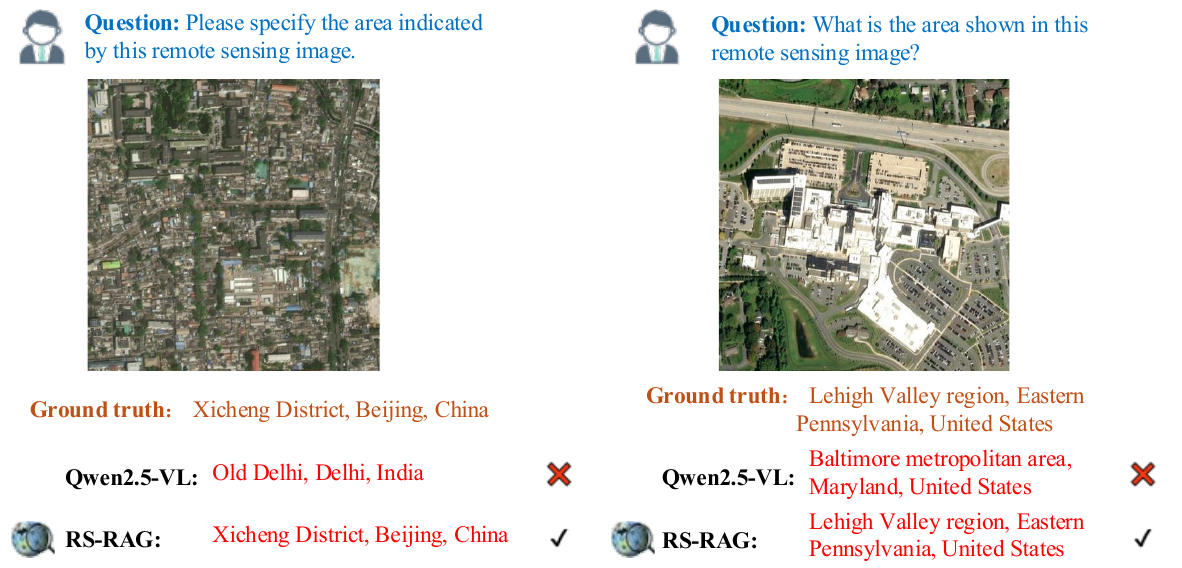}
    \caption{Qualitative results of baseline models and our RS-RAG model on the VQA task using the RSWK dataset.}
    \label{fig:urban_block_case}
\end{figure}

\rrevise{In addition to the quantitative comparison on grassland and farmland, we further collected 1,000 urban-block samples and prompted the models to identify their corresponding geographic regions. As shown in Fig.~\ref{fig:urban_block_case}, RS-RAG effectively recognizes subtle urban contexts by integrating retrieved geographic and environmental knowledge, whereas the Qwen2.5-VL often confuses visually similar structures. For instance, the baseline misclassifies \textit{Xicheng District, Beijing} as \textit{Old Delhi, India} due to architectural resemblance, while RS-RAG correctly infers the location using contextual cues such as coordinates and population density. A similar improvement is observed when distinguishing \textit{Lehigh Valley region, Pennsylvania} from \textit{Baltimore metropolitan area}, where RS-RAG exhibits more precise spatial reasoning grounded in external knowledge. These results demonstrate that RS-RAG offers clear advantages in fine-grained, visually ambiguous, and common-scene reasoning tasks, underscoring the necessity of retrieval augmentation beyond traditional visual recognition.}

\section{Conclusion}
\revise{In this work, we presented RS-RAG, a retrieval-augmented vision-language framework designed to bridge remote sensing imagery with structured domain and world knowledge, thereby addressing the challenges of knowledge-intensive multimodal reasoning. To support this framework, we constructed the RSWK dataset, a large-scale, high-resolution multimodal benchmark that includes imagery-text pairs from over 14,000 geographically and semantically diverse landmarks, spanning 16 categories across 184 countries. Building upon RSWK, we established a comprehensive benchmark covering three representative tasks, including image captioning, image classification, and VQA. To better exploit the dataset’s rich semantics, we further designed the VQA task into three distinct subtypes, targeting comprehensive reasoning, remote sensing knowledge understanding, and world knowledge understanding. Extensive experiments conducted on the RSWK benchmark and the external multimodal dataset demonstrate that RS-RAG consistently outperforms strong baselines across multiple tasks, with notable advantages in handling knowledge-intensive queries that require both factual accuracy and sequential reasoning. We believe that both the RSWK dataset and the RS-RAG framework establish a new foundation for advancing knowledge-grounded vision-language understanding in remote sensing and geospatial AI. Our work paves the way toward more capable, explainable, and generalizable multimodal systems for Earth observation and related applications.}

{\small
\bibliographystyle{IEEEtran}
\bibliography{egbib}
}

\begin{IEEEbiography}[{\includegraphics[width=1in,height=1.25in,clip,keepaspectratio]{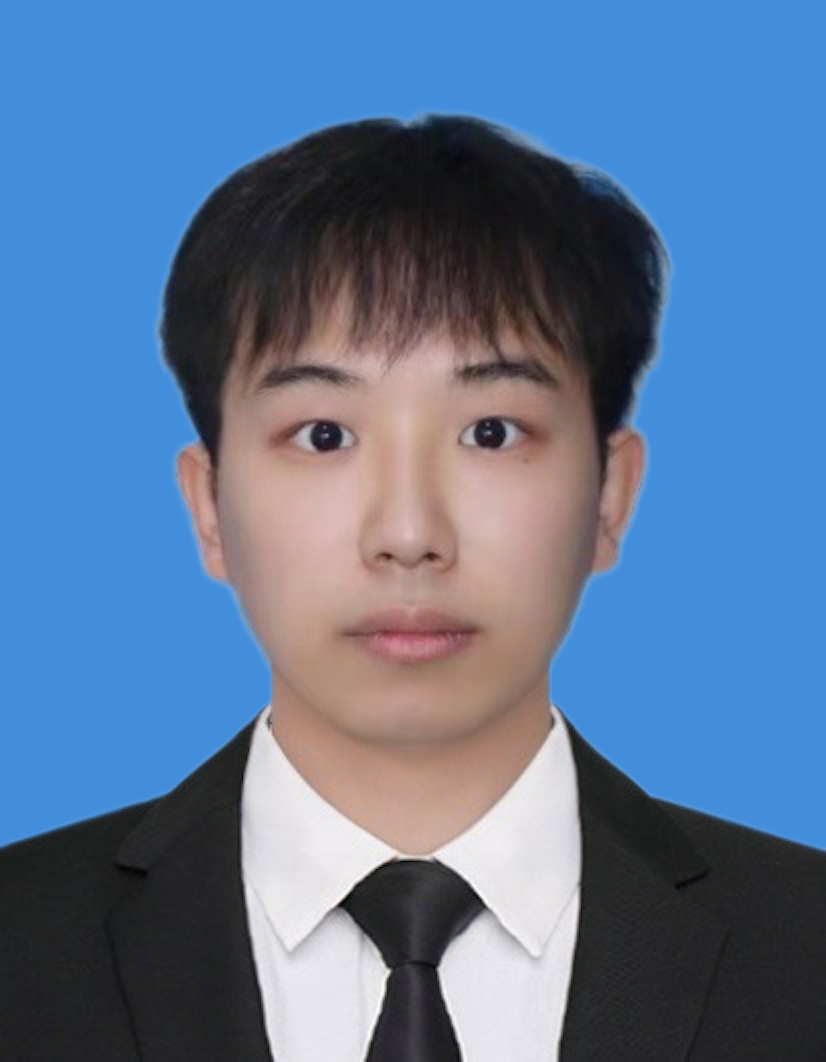}}]{Congcong Wen} (Member, IEEE) will join the China Academy of Electronics and Information Technology, Beijing, China, as a Professorate Senior Engineer (professor-equivalent) and Senior Research Scientist. He was previously a Postdoctoral Associate at New York University and New York University Abu Dhabi, and a Visiting Research Fellow at Harvard University. He received the Ph.D. degree from the Aerospace Information Research Institute, Chinese Academy of Sciences, China, in 2021. His research interests include multimodal artificial intelligence, 3D computer vision, foundation models, and remote sensing.

Dr. Wen has served in academic service roles for several well-known venues, including as a Track Chair for ICPR 2026, an Area Chair for WACV 2026, an Associate Editor of GIScience, and a Guest Editor of the Journal of Environmental \& Earth Sciences.
\end{IEEEbiography}

\begin{IEEEbiography}[{\includegraphics[width=1in,height=1.25in,clip,keepaspectratio]{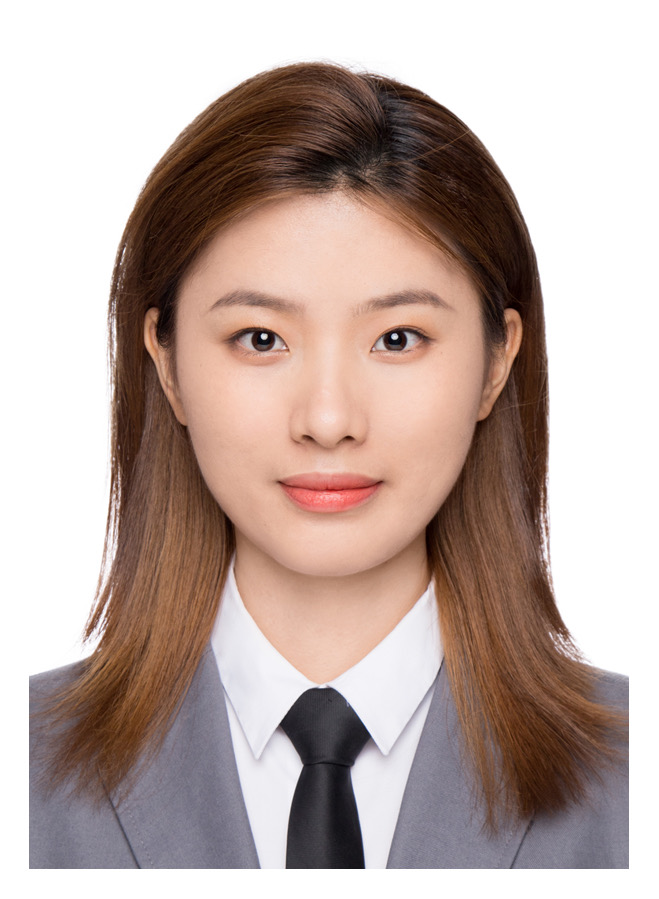}}]{Yiting Lin} is a Master's student at the School of Cyber Science and Technology, University of Science and Technology of China, No. 100 Fuxing Road, Shushan District, Hefei, Anhui, China. Major in Network and Information Security, Her research interests include Vision Large Model, Deep learning, and Retrieval-Augmented Generation applications.
\end{IEEEbiography}

\begin{IEEEbiography}[{\includegraphics[width=1in,height=1.25in,clip,keepaspectratio]{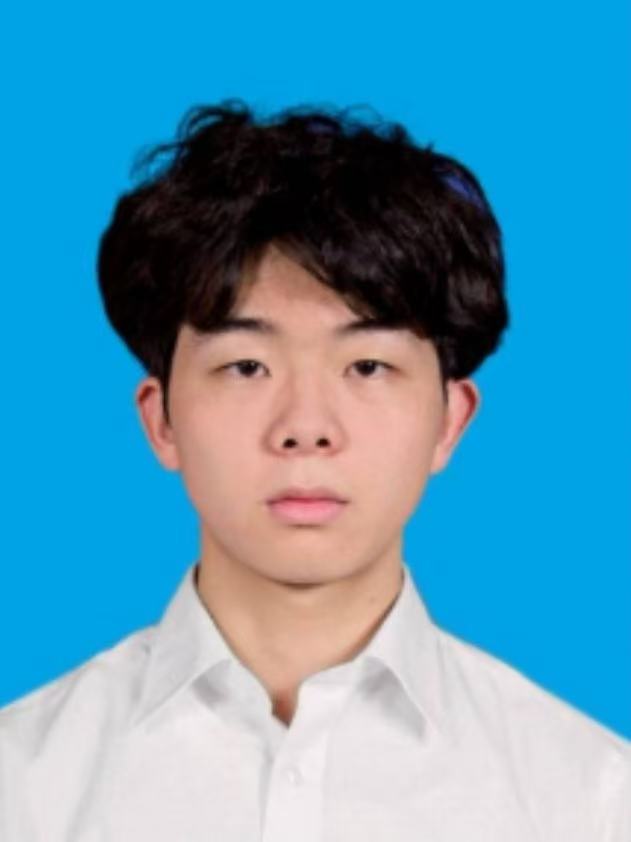}}]{Xiaokang Qu} is a Master's student at the School of Cyber Science and Technology, University of Science and Technology of China, No. 100 Fuxing Road, Shushan District, Hefei, Anhui, China. Major in Network and Information Security, His research interests include Complex Network, Network Robustness Evaluation, and Deep learning.
\end{IEEEbiography}

\begin{IEEEbiography}[{\includegraphics[width=1in,height=1.25in,clip,keepaspectratio]{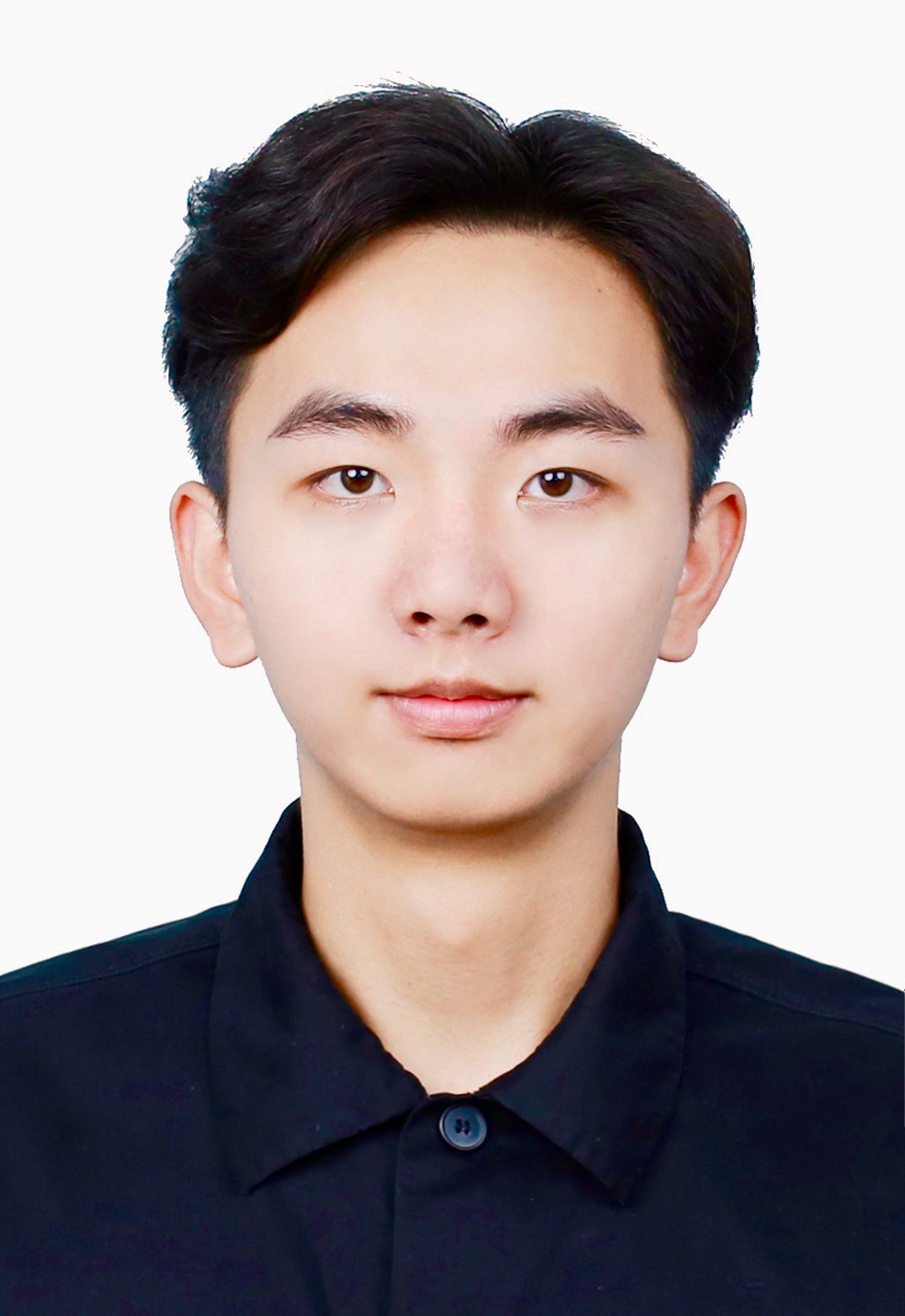}}]{Nan Li} received the Bachelor's degree in Electronic Information Science and Technology from Lanzhou University, Lanzhou, China, in 2018, and the Master degree in Data Science, City University of Hong Kong, in 2020. He is currently a Senior Engineer in China Academy of Electronics and Information Technology, Beijing, China. His research interests include multimodal large language models, reinforcement learning, The application of knowledge graphs in the field of remote sensing.
\end{IEEEbiography}

\begin{IEEEbiography}[{\includegraphics[width=1in,height=1.25in,clip,keepaspectratio]{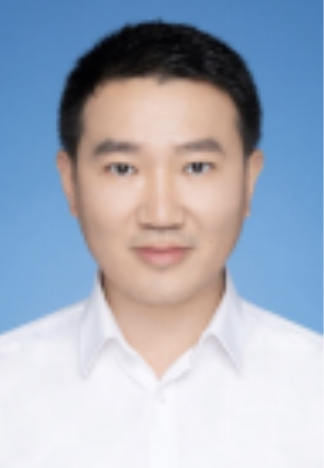}}]{Yong Liao} received the Master's degree in Computer Application Technology from the Institute of Software, Chinese Academy of Sciences, Beijing, China, in 2004, and the Ph.D. degree in Computer and Systems Engineering from the Department of Electrical and Computer Engineering, University of Massachusetts Amherst, Amherst, MA, USA, in 2010. His current research interests include cyberspace security and big data.
\end{IEEEbiography}

\begin{IEEEbiography}[{\includegraphics[width=1in,height=1.25in,clip,keepaspectratio]{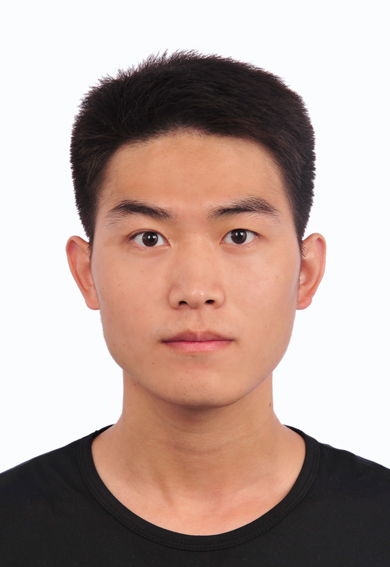}}]
{Xiang Li} is a Lecturer (Assistant Professor) in the School of Computer Science at the University of Bristol. Prior to joining the University of Bristol, he was a Lecturer at the University of Reading. He previously held postdoctoral research positions at King Abdullah University of Science and Technology (KAUST) and at New York University (NYU). His research interests include multimodal large language models, computer vision, and AI for Earth Observation. He has published over 50 papers in leading conferences and journals, including CVPR, ICCV, NeurIPS, IEEE TVCG, and IEEE TGRS. Dr. Li has served as a Guest Editor for several well-known journals, including IEEE GRSM and IEEE JSTARS. He is the recipient of the IEEE GRSS Early Career Award 2025 for his high-impact research on computer vision methods for remote sensing applications.
\end{IEEEbiography}

\begin{IEEEbiography}[{\includegraphics[width=1in,height=1.25in,clip,keepaspectratio]{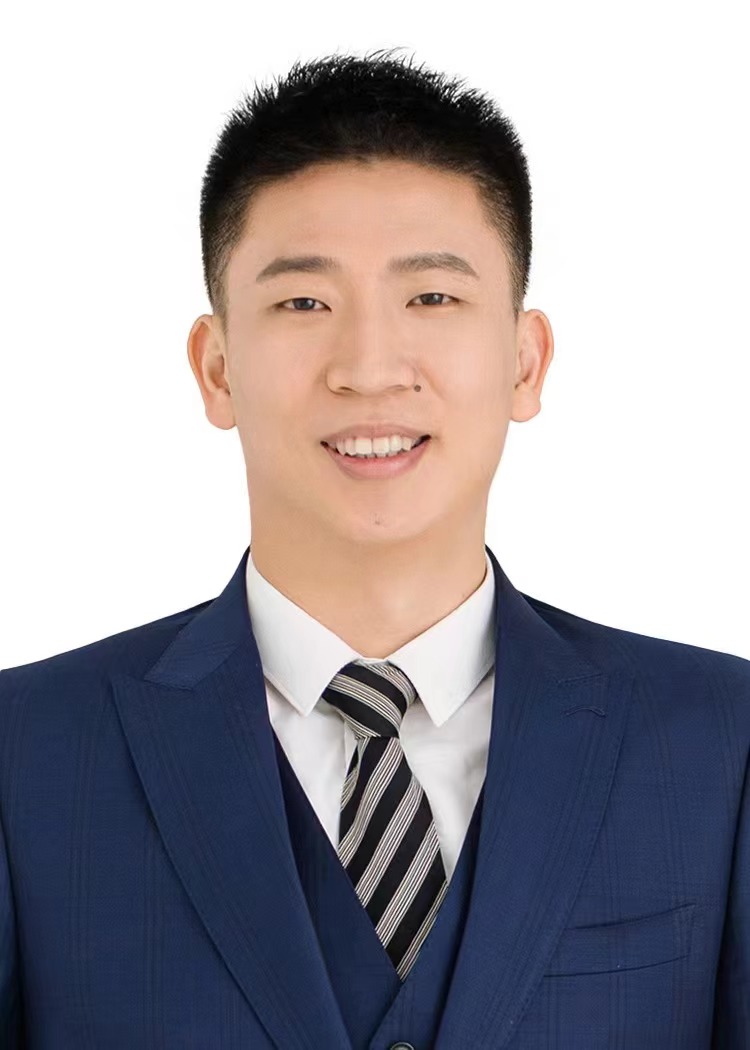}}]{Hui Lin} received the Bachelor's degree in Space Information and Digital Technology from Wuhan University, Wuhan, China, in 2012, and the Ph.D. degree in Cartography and Geographic Information Systems from the Institute of Remote Sensing and Digital Earth, Chinese Academy of Sciences, in 2017. He is currently a Senior Engineer in China Academy of Electronics and Information Technology, Beijing, China. His research interests include multimodal data fusion, deep learning, and remote sensing applications.
\end{IEEEbiography}

\end{document}